\documentclass[twoside,11pt]{article}
\usepackage{fullpage}
\usepackage{times}
\usepackage{capt-of}

\usepackage[numbers]{natbib}
\usepackage{multicol}
\usepackage[bookmarks=true]{hyperref}

\usepackage{graphicx}
\usepackage{subcaption}
\usepackage{amsmath,amssymb,mathtools}
\usepackage{array}
\usepackage{url}
\usepackage{gensymb}

\usepackage{xcolor}

\usepackage{enumitem}

\usepackage{algorithmic}
\usepackage[linesnumbered,lined,ruled]{algorithm2e}
\SetAlCapSty{}

\SetCommentSty{mycommfont}

\usepackage{breakurl}

\newcommand{\stt}[1]{{\small\texttt{#1}}}

\newtheorem{example2}{\bf Example}
\newtheorem{execexample}{\bf Execution Example}

\newcounter{ctr}

\begin{document}

\title{Combining Commonsense Reasoning and Knowledge Acquisition to
  Guide Deep Learning in Robotics
  % \thanks{Grants or other notes about the article that should go on
  %   the front page should be placed here. General acknowledgments
  %   should be placed at the end of the article.}
}

\author{  Mohan Sridharan\\
  School of Computer Science, University of Birmingham, UK\\
  Tel.: +44-121-414-2793, Fax: +44-121-414-4281\\
  \texttt{m.sridharan@bham.ac.uk} % \\
  \and
  Tiago Mota\\
  Electrical and Computer Engineering, The University of Auckland, NZ \\
  Tel.: +64-9-373-7599, Fax: +64-9-373-4281 \\
  \texttt{tmot987@aucklanduni.ac.nz} % \\
}

%\date{Received: date / Accepted: date}
% The correct dates will be entered by the editor

\maketitle

%%%%%%%%%%%%%%%%%%%%%%%%%%%%%%%%%%%%%%%%%%%%%%%%%%%%%%%%%%%%%%%%%%%%%%%%%%%%%%%%%%%%
%%%%%%%%%%%%%%%%%%%%%%%%%%%%%%%%%%%%%%%%%%%%%%%%%%%%%%%%%%%%%%%%%%%%%%%%%%%%%%%%%%%%
\begin{abstract}
  Algorithms based on deep network models are being used for many
  pattern recognition and decision-making tasks in robotics and AI.
  Training these models requires a large labeled dataset and
  considerable computational resources, which are not readily
  available in many domains. Also, it is difficult to explore the
  internal representations and reasoning mechanisms of these models.
  As a step towards addressing the underlying knowledge
  representation, reasoning, and learning challenges, the architecture
  described in this paper draws inspiration from research in cognitive
  systems. As a motivating example, we consider an assistive robot
  trying to reduce clutter in any given scene by reasoning about the
  occlusion of objects and stability of object configurations in an
  image of the scene. In this context, our architecture incrementally
  learns and revises a grounding of the spatial relations between
  objects and uses this grounding to extract spatial information from
  input images.  Non-monotonic logical reasoning with this information
  and incomplete commonsense domain knowledge is used to make
  decisions about stability and occlusion.  For images that cannot be
  processed by such reasoning, regions relevant to the tasks at hand
  are automatically identified and used to train deep network models
  to make the desired decisions. Image regions used to train the deep
  networks are also used to incrementally acquire previously unknown
  state constraints that are merged with the existing knowledge for
  subsequent reasoning. Experimental evaluation performed using
  simulated and real-world images indicates that in comparison with
  baselines based just on deep networks, our architecture improves
  reliability of decision making and reduces the effort involved in
  training data-driven deep network models.  

  % \keywords{Non-monotonic
%     logical reasoning \and Deep learning \and Decision tree induction
%     \and Scene understanding}
\end{abstract}

%%%%%%%%%%%%%%%%%%%%%%%%%%%%%%%%%%%%%%%%%%%%%%%%%%%%%%%%%%%%%%%%%%%%%%%%%%%%%%%%%%%%
%%%%%%%%%%%%%%%%%%%%%%%%%%%%%%%%%%%%%%%%%%%%%%%%%%%%%%%%%%%%%%%%%%%%%%%%%%%%%%%%%%%%
\section{Introduction}
\label{sec:introduction}
\begin{figure}[tb]
  \begin{center}
    \includegraphics[height=2in]{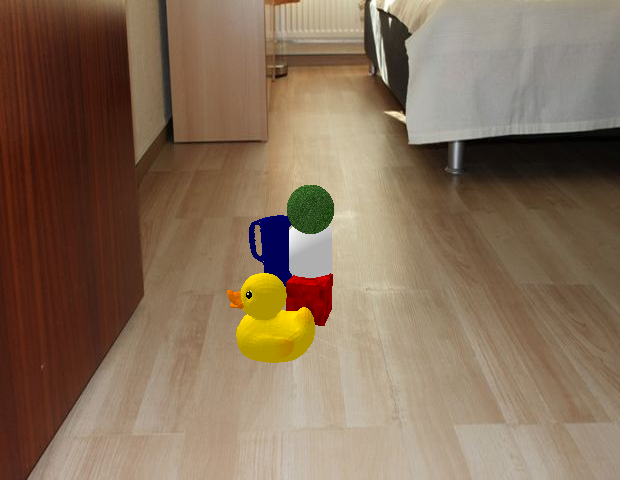}
    \caption{A simulated scene with toys on the ground. The robot has
      to reason about occlusion of objects and stability of object
      configurations to reduce clutter.}
    \label{fig:illus-image}
    \vspace{-2em}
  \end{center}
\end{figure}

Imagine an assistive robot\footnote{We use the terms ``robot'',
  ``learner'', and ``agent'' interchangeably in this paper.}  that has
to clear away toys and objects that children have (mis)placed in
different configurations in different rooms of a home.
Figure~\ref{fig:illus-image} shows an example simulated scene of toys
in a room. The robot's task poses knowledge representation, reasoning,
and learning challenges because it is difficult to provide labeled
training examples of all possible arrangements of objects or of all
combinations of object attributes.  Also, the robot has to reason with
different descriptions of incomplete domain knowledge and the
associated uncertainty. This may include qualitative descriptions of
commonsense knowledge about the domain objects, e.g., \emph{default}
statements such as ``structures with a large object placed on a small
object are typically unstable'' that hold in all but a few exceptional
circumstances, and an initial grounding of the relations between
objects, e.g., meaning in the physical world for ``left of'' and
``behind''. The robot may also obtain quantitative descriptions of
knowledge and uncertainty, e.g., from algorithms for sensing and
navigation that provide information such as ``I am 90\% certain the
big red box is stable''.  Furthermore, human participants are unlikely
to have the time and expertise to interpret raw sensor data or to
provide comprehensive feedback, and the robots' decisions may be
incorrect or sub-optimal if they are based on incomplete or incorrect
domain knowledge.

We use the robot assistant (RA) domain described above as a motivating
example to describe the capabilities of an architecture that is a step
towards addressing the underlying knowledge representation, reasoning,
and learning challenges. In particular, we consider the scene
understanding tasks of estimating the \emph{partial occlusion} of
objects and the \emph{stability} of object configurations. State of
the art methods for such tasks are based on deep network
architectures. Although these methods often provide high accuracy,
they require a large number of labeled training examples, are
computationally expensive, and make it difficult to understand the
decisions made. Research in cognitive systems has demonstrated the
benefits of considering different representations and reasoning
methods, and of coupling representation, reasoning, and learning such
that they inform and guide each other.  Drawing inspiration from such
systems, our architecture exploits the complementary strengths of
non-monotonic logical reasoning based on incomplete commonsense domain
knowledge, deep learning, and incremental inductive learning of
constraints governing domain states.

In the context of the RA domain, a robot equipped with our
architecture obtains 3D point clouds of the scenes of interest.  The
robot is also provided incomplete domain knowledge in the form of the
type and attributes (e.g., color, size) of some domain objects; an
initial qualitative grounding (i.e., meaning in the physical world) of
prepositional words (e.g., ``above'', ``left\_of'', and ``near'')
describing spatial relations between objects in the scene; and some
axioms (i.e., rules) governing actions, states, and change in the
domain, including default statements such as ``any object
configuration with a large object placed on a small object is
typically unstable'' that is true in all but a few exceptional
circumstances. The architecture enables the robot to:
\begin{enumerate}[label=(\arabic*)]
\item Interactively and cumulatively learn and revise a quantitative
  (i.e., metric) grounding of the spatial relations between objects,
  using the qualitative grounding, point cloud data of objects in a
  small number of input (RGB-D) images, and limited feedback from
  humans.

\item Use non-monotonic logical reasoning to perform the estimation
  tasks on any given input image using a relational representation of
  the information extracted from the image (e.g., object attributes,
  spatial relations between objects) and the incomplete (prior) domain
  knowledge.

\item Automatically identify regions of interest (ROIs) in images for
  which the estimation tasks cannot be completed using non-monotonic
  logical reasoning. These ROIs and the corresponding (occlusion,
  stability) labels are used to guide the construction of deep
  networks during training, and the ROIs are processed by the trained
  networks during testing, for the estimation tasks.

\item Incrementally learn previously unknown relational constraints
  using the information used to train the deep networks, i.e., the
  relational information extracted from the ROIs and the corresponding
  labels. A heuristic approach inspired by human forgetting helps
  merge and update the learned and existing knowledge for subsequent
  reasoning.
\end{enumerate}
We use CR-Prolog, an extension of Answer Set Prolog
(ASP)~\cite{balduccini:aaaisymp03,gelfond:aibook14}, for non-monotonic
logical reasoning with incomplete commonsense domain
knowledge\footnote{We will use the terms ``ASP'' and ``CR-Prolog''
  interchangeably in this paper.}. We also adapt existing network
architectures for the deep learning component of our architecture. The
design and evaluation of our architecture is \emph{subject to two
  caveats}:
\begin{itemize}[label=$\star$]
\item State of the art scene understanding methods often focus on
  generalizing across different tasks and domains using a large number
  of labeled training examples. Our focus, on the other hand, is on
  enabling a robot to use a small number of images and to
  automatically limit learning to previously unknown information
  relevant to the tasks at hand in any given domain. We thus do not
  use existing benchmark datasets for experimental comparison.
  Instead, we use a small set of real world and simulated images of
  scenes in the robot assistant domain described above.
  % ---Figure~\ref{fig:illus-image} shows an example simulated scene.

\item Given our focus on exploring the interplay between
  representation, reasoning, and learning, we do not compare our
  architecture's performance against existing methods for just the
  reasoning or learning components. Instead, we run ablation studies
  and compare against data-driven baselines that use deep networks to
  perform the estimation tasks of interest. For ease of understanding,
  we also limit perceptual processing to that of 3D point clouds of
  scenes, and limit previously unknown domain knowledge to state
  constraints.  We have explored the learning of other types of
  axioms, and the use of these axioms to interpret the behavior of
  deep networks, in other work~\cite{mota:SNCS21,mohan:ACS18}.
\end{itemize}
Some components of this architecture have been described in conference
papers, e.g., the incremental learning of a quantitative grounding of
spatial relations~\cite{mota:ijcai18} and the incremental learning of
state constraints~\cite{mota:rss19,riley:FRONTIERS19}. Here, we
describe these components in detail, highlight recent revisions to
these components, and explore the capabilities supported by the
interplay between these components. We also provide additional
experimental results to demonstrate a marked increase in the accuracy
of decision-making and a reduction in the computational effort in
comparison with architectures that only use deep networks.

The remainder of this paper is organized as follows.  First,
Section~\ref{sec:relwork} discusses related work to motivate our
architecture, whose components are described in
Section~\ref{sec:arch}. The experimental setup and results are
described in Section~\ref{sec:expres}. Finally, the conclusions and
directions for further research are discussed in
Section~\ref{sec:conclusions}.

%%%%%%%%%%%%%%%%%%%%%%%%%%%%%%%%%%%%%%%%%%%%%%%%%%%%%%%%%%%%%%%%%%%%%%%%%%%%%%%%%%%%
%%%%%%%%%%%%%%%%%%%%%%%%%%%%%%%%%%%%%%%%%%%%%%%%%%%%%%%%%%%%%%%%%%%%%%%%%%%%%%%%%%%%
\section{Related Work}
\label{sec:relwork}
The scene understanding tasks in the motivating RA domain considered
in this paper are representative of a wide range of estimation and
prediction problems that pose the knowledge representation, reasoning,
and learning problems of interest. Deep networks provide state of the
art performance for these problems and for many other computer vision
and control problems. For instance, a Convolutional Neural Network
(CNN) has been used to predict the stability of a tower of
blocks~\cite{lerer2016,li2016}, the movement of an object sliding down
an inclined surface and colliding with another object~\cite{wu2015},
and the trajectory of an object after bouncing against a
surface~\cite{purushwalkam2019}. However, CNNs and other deep networks
require a large number of labeled examples and considerable
computational resources to learn the mapping from inputs to outputs.
In addition, it is difficult to understand the operation of the
learned networks, which also makes it challenging to transfer
knowledge learned in one scenario or task to a related scenario or
task~\cite{sunderhauf2018,zhang2016}.

Since labeled training examples are not readily available in many
domains, researchers have explored approaches that simulate labeled
data or use prior knowledge to constrain learning. For instance,
physics engines have been used to generate labeled data for training
deep networks that predict the movement of objects in response to
external
forces~\cite{fragkiadaki2015,mottaghi2016,wagner:eccvwrkshp18}, or for
understanding the physics of scenes~\cite{battaglia2013}. A recurrent
neural network (RNN) architecture augmented by arithmetic and logical
operations has been used to answer questions about
scenes~\cite{neelakantan2015}, but it used textual information instead
of the more informative visual data and did not support reasoning with
commonsense knowledge. Another example is the use of prior knowledge
to encode state constraints in the CNN loss function; this reduces the
effort in labeling training images, but it requires the constraints to
be encoded manually as loss functions for each
task~\cite{stewart2017}. The structure of deep networks has also been
used to constrain learning, e.g., by using relational frameworks for
visual question answering (VQA) that consider pairs of objects and
related questions to learn the relations between
objects~\cite{santoro2017}. This approach, however, only makes limited
use of the available knowledge, and does not revise the constraints
over time.

For many problems in robotics and AI, prior domain knowledge often
includes the grounding, i.e., an interpretation in the physical world,
of words such as \emph{in}, \emph{behind}, and \emph{above}
representing spatial relations between objects. Many systems for
grounding such relations are based on manually encoded descriptions,
or on learning algorithms. Methods using the former typically rely on
Qualitative Spatial Representations (QSR) of spatial
relations~\cite{elliott2015,ye2013,zampogiannis2015}, whereas in the
latter case, it is more common to use Metric Spatial Representations
(MSR), i.e., measures based on distances and
angles~\cite{belz2015,mees2017}. QSR-based approaches often
approximate objects as points and establish static boundaries between
spatial relations, whereas the grounding of spatial relations is
likely to change over time in dynamic domains.  MSR-based systems
often learn the grounding of spatial relations offline or in a
separate training phase. Specific instances of QSR and MSR have been
used for different tasks in robotics and computer vision, e.g., QSR
relations have been extracted from videos~\cite{gatsoulis2016}, MSR
and kd-trees have been used to infer spatial relations between
objects~\cite{ziaeetabar2017}, QSR and MSR have been compared for
scene understanding on robots~\cite{thippur2015}, the relative
position of objects has been used to predict successful action
execution~\cite{fichtl2015}, and methods have been developed to reason
about and learn spatial relations between
objects~\cite{jund2018,krishnaswamy2019}. Specialized meetings have
explored the use of natural language to describe spatial relationships
between objects~\cite{dobnik2018,ulinski2019}. Deep networks have also
been used to infer spatial relations between objects using images and
natural language expressions, for manipulation~\cite{paul2018},
navigation~\cite{pronobis2017}, and human-robot
interaction~\cite{shridhar2017}. Our approach for learning the
grounding of spatial relations starts with a manually-encoded generic
(QSR) grounding, and interactively learns a specialized (MSR)
grounding from experience and human feedback~\cite{mota:ijcai18}.

There is an established literature on generic methods for learning
domain knowledge. Examples include the incremental revision of a
first-order logic representation of action
operators~\cite{gil:icml94}, the use of inductive logic programming to
learn domain knowledge represented as an Answer Set Prolog (ASP)
program~\cite{law:AIJ18,law:ALP20}, and work in our group on coupling
of non-monotonic logical reasoning, inductive learning, and relational
reinforcement learning to incrementally acquire actions and
axioms~\cite{mohan:ACS18}. Our approach for learning domain axioms is
inspired by work in interactive task learning, a general framework for
acquiring domain knowledge using labeled examples or reinforcement
signals obtained from domain observations, demonstrations, or human
instructions~\cite{chai:ijcai18,laird:IS17}. However, unlike
approaches that learn from many training examples, our approach seeks
to incrementally acquire and revise domain knowledge from limited,
partially-defined, training examples. It can be viewed as building on
early work on heuristic search through the space of hypotheses and
observations~\cite{simon:bookchap-kc74}, but such methods have rarely
been explored for scene understanding.

Studies in neural-symbolic learning and reasoning have explored the
benefits and limitations of interleaving statistical learning and
symbolic reasoning~\cite{besold2017,samek:ITU17}. For example, the
probabilistic logic programming language ProbLog has been extended to
DeepProbLog, which supports symbolic and sub-symbolic inference,
program induction, and deep (neural) learning from examples based on
neural predicates~\cite{manheave:nips18}. Another example is a
neural-symbolic visual question answering system that uses deep
networks to infer structural object-based scene representation from
images, and to generate a hierarchical (symbolic) program of
functional modules from the question. Running the program on the
representation answers the desired question~\cite{yi:nips18}. Many of
these methods rely on classical first-order logic~\cite{guillame2010}
that is not expressive enough for reasoning with commonsense
knowledge, use simplified neural architectures corresponding to
specific symbolic representations~\cite{garcez2007}, associate
probability values with all logic statements which is not always
meaningful, or do not clearly establish the link between reasoning and
learning. In addition, although retracting imperfect or incorrect
beliefs has long been considered as important as learning new
knowledge~\cite{charniak1978,granger1980}, existing neural-symbolic
approaches rarely support the automatic detection and correction of
errors in learned knowledge. In parallel, there has been much work on
interpreting the operation of deep networks, e.g., by computing
gradients and decompositions at different layers of the network, and
providing heatmaps that indicate the features most relevant to the
observed output(s) of deep network~\cite{assaf:ijcai19,samek:ITU17}.
However, these approaches do not exploit the incomplete commonsense
domain knowledge for reliable and efficient reasoning and learning,
and for generating explanations in the form of relational
descriptions.

% I found the list of papers for scene understanding to be not very recent. Publications are dated prior 2018, with the exception of very few references dates 2018-2019. Although not directly related to scene understanding there are quite few approaches and architecture that are used today to extract information from images, video and scenes. It would be nice to discuss why any of the approaches (e.g. http://clevrer.csail.mit.edu) are not applicable or differ from the method proposed here.
% CLEVRER
% clevrer.csail.mit.edu
% CLEVRER: CoLlision Events for Video REpresentation and Reasoning

A recent evaluation of state of the art computational models for
reasoning (and understanding) on a diagnostic video dataset revealed
that existing methods are able to answer descriptive questions about
the scene, but perform poorly on explanatory, predictive, and
counterfactual questions~\cite{yi:iclr20}. The results indicate that
causal reasoning methods need an understanding of the domain dynamics
and causal relations. This understanding can be provided using models
of domain physics, as described above. Work in our research group, on
the other hand, is inspired by research in cognitive systems, which
indicates that coupling knowledge representation, reasoning, and
interactive learning, can help address the limitations described
above~\cite{gomez:AMAI21,mota:SNCS21,riley:FRONTIERS19,mohan:ACS18,mohan:JAIR19}.
The architecture described in this paper combines the complementary
strengths of deep learning, reasoning with incomplete commonsense
knowledge, and inductive learning. It explores the interplay between
our prior work on incrementally learning a grounding of spatial
relations between objects~\cite{mota:ijcai18} and on learning
constraints that govern domain
states~\cite{mota:rss19,riley:FRONTIERS19}, introduces a heuristic
approach inspired by human forgetting~\cite{ellwart2019} to detect and
correct errors while merging the learned knowledge with existing
knowledge, and describes detailed results of ablation studies and
other experiments evaluating the capabilities of our architecture.

%%%%%%%%%%%%%%%%%%%%%%%%%%%%%%%%%%%%%%%%%%%%%%%%%%%%%%%%%%%%%%%%%%%%%%%%%%%%%%%%%%%%
%%%%%%%%%%%%%%%%%%%%%%%%%%%%%%%%%%%%%%%%%%%%%%%%%%%%%%%%%%%%%%%%%%%%%%%%%%%%%%%%%%%%
\section{Reasoning and Learning Architecture}
\label{sec:arch}
\begin{figure*}[tb]
  \begin{center}
    \includegraphics[width=0.95\linewidth]{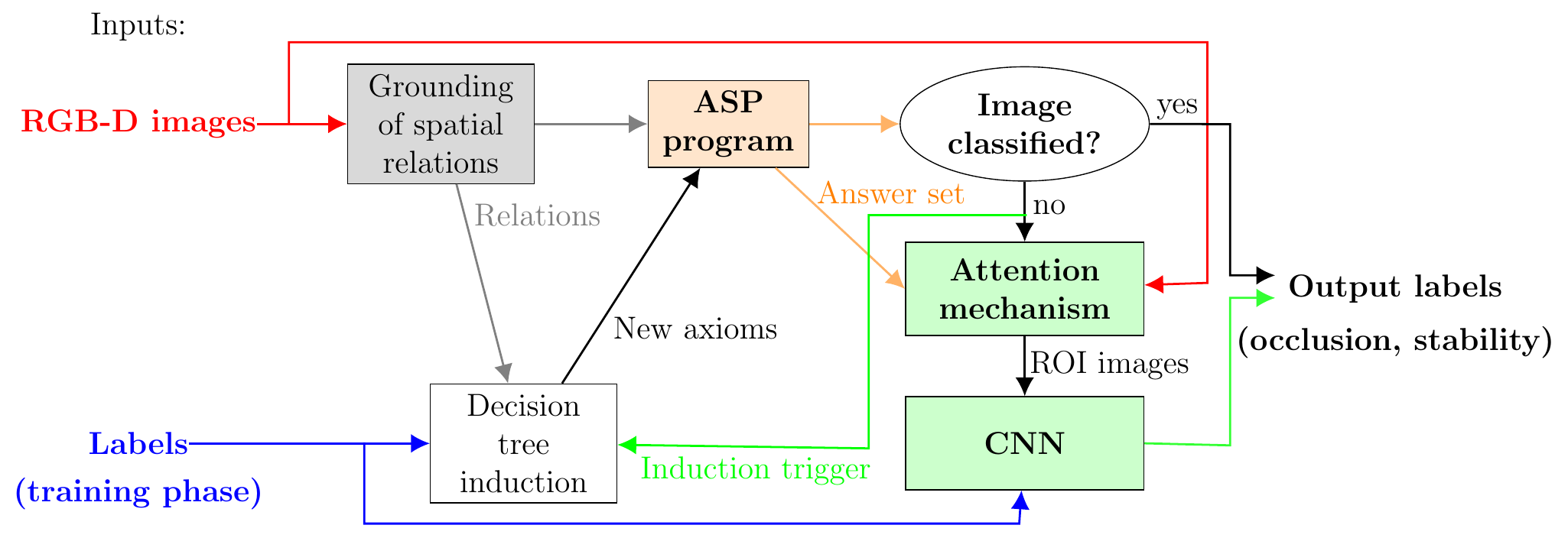}
    \caption{Architecture combines the complementary strengths of deep
      learning, non-monotonic logical reasoning with incomplete
      commonsense domain knowledge, and decision tree induction;
      architecture illustrated here in the context of estimating
      occlusion of objects and stability of object structures.}
    \label{fig:architecture}
    \vspace{-1em}
  \end{center}
\end{figure*}

Figure~\ref{fig:architecture} is an overview of our reasoning and
learning architecture whose components are adapted and described in
this paper in the context of estimating the occlusion of objects and
stability of object configurations in the RA domain. An object is
considered to be occluded if the view of any minimal fraction of its
frontal face is hidden by another object, and a configuration (or
structure), i.e., a stack of objects, is unstable if any object in the
structure is unstable. This architecture takes as input RGB-D images
of scenes with different object configurations.  During training, the
inputs also include the occlusion labels of objects and the stability
labels of object configurations in these images. Also, a static
qualitative representation is provided for the grounding of the
prepositional words encoding spatial relations between objects; this
grounding is used to learn and revise a histogram-based metric
representation of the grounding.

For any given image, our architecture first attempts to assign the
desired (occlusion and stability) labels to scene objects using
ASP-based non-monotonic logical reasoning. This reasoning considers
the incomplete commonsense domain knowledge and the relational (i.e.,
logic-based) representation of the noisy information extracted from
the RGB-D image, i.e., object attributes and spatial relations between
objects. If such reasoning is able to complete the estimation tasks,
i.e., provide correct labels during training and estimate labels
during testing, no further analysis of this image is performed.
Otherwise, an ``\emph{attention mechanism}'' reasons with existing
knowledge (of task and domain) to \emph{automatically} identify
Regions of Interest (ROIs) in the image relevant to the tasks to be
performed, with each ROI containing one or more objects. A CNN is
trained with image ROIs and the corresponding ground truth labels, and
used (during testing) to map these ROIs to the desired labels. In
addition, ROIs used to train the CNN are also used as input to a
decision tree induction algorithm that maps object attributes and
spatial relations to the target labels. Branches in the tree that have
sufficient support among the training examples are used to construct
axioms representing state constraints.  The learned constraints are
automatically and heuristically merged with the existing domain
knowledge by adding or removing axioms as appropriate, and used for
subsequent reasoning. We will use the following illustrative domain to
describe the components of our architecture in more detail.

\begin{example2}\label{ex:illus-example}[Robot Assistant (RA) Domain]\\
  {\rm A simulated robot analyzes images of scenes containing objects
    in different configurations. The goal is to estimate occlusion of
    objects and stability of object structures, and to rearrange
    object structures so as to minimize clutter. Domain knowledge
    includes incomplete information about the object's attributes such
    as $size$ (small, medium, large), $surface$ (flat, irregular), and
    $shape$ (cube, cylinder, duck), and the spatial (i.e., geometric)
    $relation$ between objects (above, below, front, behind, right,
    left, close).  The robot can move objects to achieve the desired
    goals. Domain knowledge also includes some axioms governing domain
    dynamics (i.e., states and actions) but some other axioms may be
    unknown, e.g.:
    \begin{itemize}
    \item Placing an object on top of an object with an irregular
      surface causes instability;
    \item An object is not occluded if all objects in front of it are
      moved away.
    \end{itemize}
    In addition to learning the previously unknown axioms, the
    existing axioms may also need to be revised over time based on
    observations, e.g., the robot may find that it is possible to
    place an object on another with an irregular surface under certain
    conditions.  }
\end{example2}
Using this domain as a running example, Section~\ref{sec:arch-ground}
first describes the approach for incremental and interactive grounding
of the spatial relations between objects.  Section~\ref{sec:arch-kr}
describes the approach for representing and reasoning with incomplete
commonsense domain knowledge and the information extracted from images
to assign the desired labels. Next, Section~\ref{sec:arch-roicnn}
describes the identification of the relevant ROIs in images for which
ASP-based reasoning could not assign (correct) labels, and the
learning of CNNs that map features from these ROIs to the desired
labels. Finally, Section~\ref{sec:arch-dtmerge} describes the
incremental decision-tree induction of previously unknown domain
axioms, and the heuristic approach to merge and revise the new axioms
and existing knowledge.

%%%%%%%%%%%%%%%%%%%%%%%%%%%%%%%%%%%%%%%%%%%%%%%%%%%%%%%%%%%%%%%%%%%%%%%%%%%%%%%%%%%%

\subsection{Grounding of Spatial Relations}
\label{sec:arch-ground}
\begin{figure}[tb]
  \centering
  \includegraphics[width=.5\textwidth,height=4.5cm]{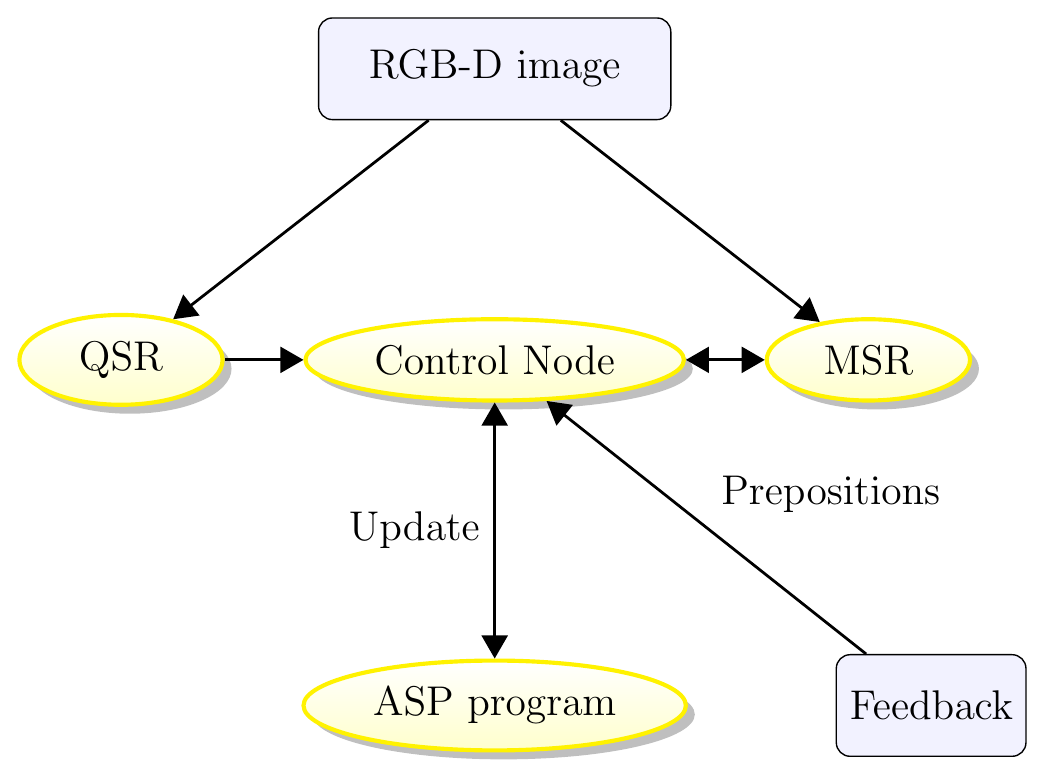}
  \caption{Overview of approach for grounding spatial relations
    between objects. A static QSR-based grounding and human input (if
    available) are used to incrementally learn and revise a MSR-based
    grounding.}
  \label{fig:grounding}
  \vspace{-1em}
\end{figure}

Our architecture includes a hybrid approach, which combines a
Qualitative Spatial Representation (QSR) and a Metric Spatial
Representation (MSR), for grounding (i.e., assigning meaning in the
physical world for) the prepositional words encoding the spatial
relations between scene objects. Figure~\ref{fig:grounding} presents
an overview of our approach. We consider seven position-based
prepositions (\textit{in, above, below, front, behind, right, left})
and three distance-based prepositions (\textit{touching, not-touching,
  far}).  These prepositions are used to encode spatial relations
between specific scene objects as logic statements in an ASP program.
The QSR module provides an initial, manually-encoded, generic
grounding of spatial relations, which is used to extract spatial
relations between pairs of 3D point clouds in any given input scene.
Human feedback, when available, is in the form of labels for the
spatial relations between pair(s) of point clouds in a scene. Both the
QSR-based output and human feedback are transmitted by the control
node to the MSR module, which incrementally acquires and revises a
histogram-based grounding of the prepositions for spatial relations.
Assuming human feedback to be accurate, the control node also computes
the relative trust in the two groundings (QSR, MSR). The more reliable
grounding is used to extract spatial relations between scene objects
in subsequent images. This information is translated to logic
statements that are added to the ASP program for inference. The
individual modules of this approach are described below.

%%%%-----------------------------------------------------------------------------
\subsubsection{Qualitative Spatial Representation}
\label{sec:arch-ground-qsr}
\begin{figure}[tb]
\centering
  \begin{subfigure}[b]{.30\textwidth}
    \centering
    \includegraphics[width=\textwidth,height=3.2cm]{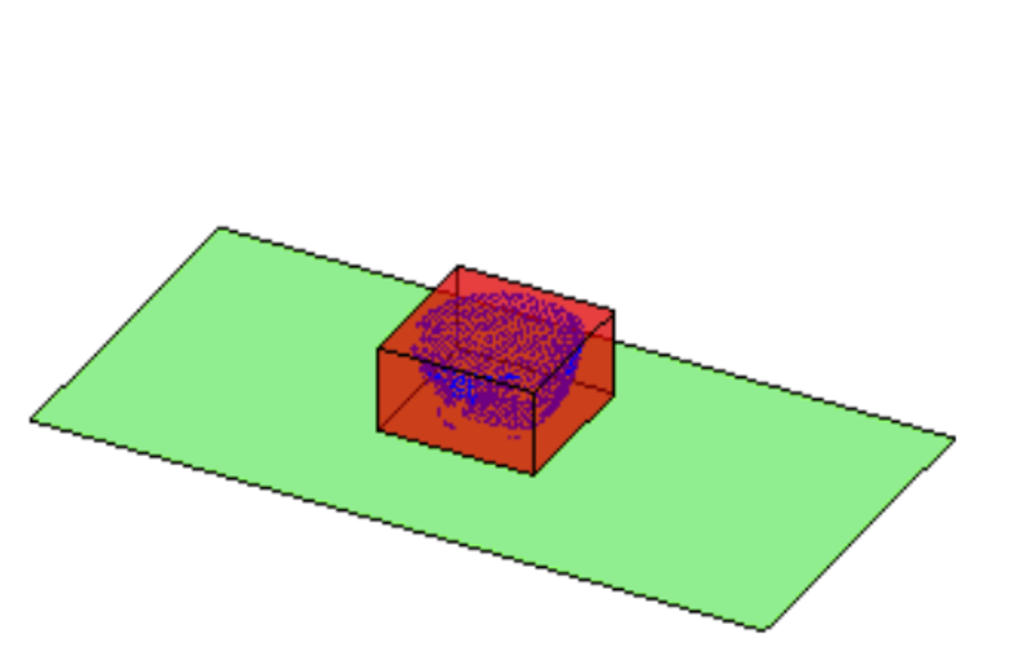}
    \caption{}%Bounding Box.}
    \label{fig:BB-init}
  \end{subfigure}
  \hspace{5.5em}
  %\hfill
  \begin{subfigure}[b]{.30\textwidth}
    \centering
    \includegraphics[width=\textwidth,height=3.2cm]{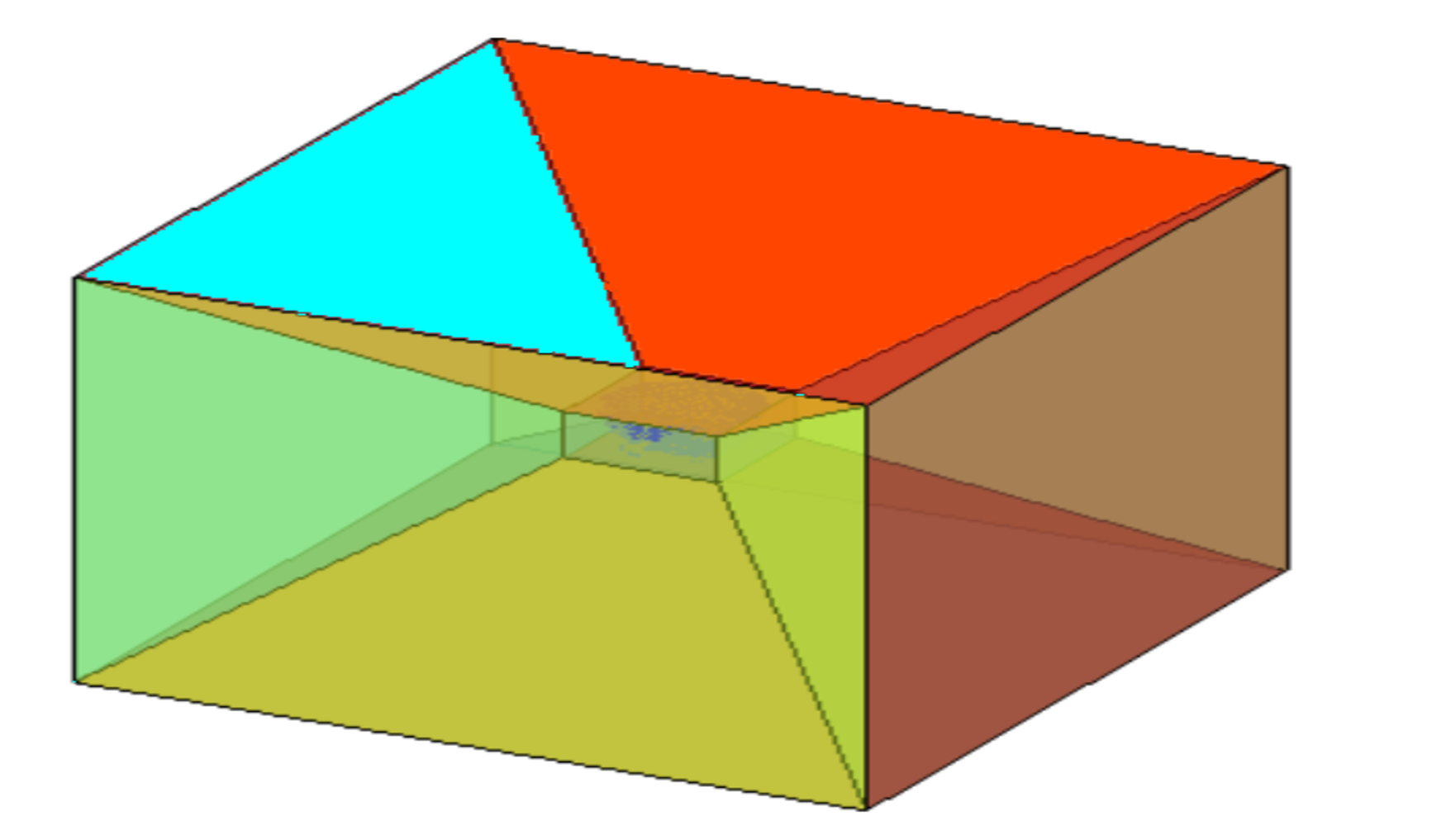}
    \caption{}%Six Pyramids.}
    \label{fig:BB-pyramids}
  \end{subfigure}
  \vspace{-0.5em}
  \caption{QSR representation: (a) Bounding box for point cloud of a
    particular object; and (b) Pyramids delimiting space around the
    bounding box.}
  \label{fig:BB}
  %\vspace{-1em}
\end{figure}

Our QSR model is based on an established approach described
in~\cite{zampogiannis2015}. For each 3D point cloud in any given
image, a bounding box containing it (i.e., a convex cuboid around the
object) is created---see Figure~\ref{fig:BB-init}.  If this point
cloud is considered the reference object, the space around this object
is divided into non-overlapping pyramids representing the relations
\textit{left, right, front, behind, above} and \textit{below}---see
Figure~\ref{fig:BB-pyramids}. In our implementation, the spatial
relation of an object with respect to a reference object is determined
by the non-overlapping pyramid around the reference that has most of
the point cloud of the object under consideration. Also, any object
with most of its point cloud located inside the bounding box of the
reference object is said to be ``\textit{in}'' the reference object.
In this paper, we disregard the fact that this definition of \emph{in}
can lead to errors, especially in domains with non-convex objects,
e.g., a book that is actually \emph{under} a large table may be
classified (incorrectly) as being \emph{in} the table because the
bounding box of the table envelopes most of the point cloud of the
book.

For ease of representation, our approach differs
from~\cite{zampogiannis2015} in the definition of the distance-related
prepositions: \textit{touching, not-touching} and \textit{far}.  For a
pair of point cloud clusters, the $10\%$ closest distances between
pairs of points drawn from the point clouds are computed, and the
following heuristics are used to determine if the two objects are
\emph{touching}, \emph{not touching}, or \emph{far} (i.e., distinctly
separated) from each other:
\begin{align}
  \label{eqn:distance}
  &touching\ \Rightarrow\ distance(10\%)\ \leq\ 0.01 \\ \nonumber &not
  \textnormal{-} touching\ \Rightarrow\ 0.01 < distance(10\%)\ < 1.0\\
  \nonumber &far\ \Rightarrow\ distance(10\%)\ \geq\ 1.0
\end{align}
where distances are measured in meters. In other words, two objects
are touching if the $10\%$ closest distances are less than or equal to
$1 cm$. The generic, manually-encoded grounding based on this QSR
model does not change over time, whereas changes in factors such as
camera pose may require the grounding to be revised. However, based on
the reasonable assumption that the robot's initial estimate of spatial
relations is based on its view of the scene, our approach has the
robot use the QSR-based grounding to identify spatial relations
between objects in the initial stages. This grounding and human input
of spatial relations between object pairs (when available) are used to
incrementally learn and revise a specialized, quantitative grounding
of spatial relations between objects, which we describe next.

%%%%-----------------------------------------------------------------------------
\subsubsection{Metric Spatial Representation}
\label{sec:arch-ground-msr}
Unlike the QSR-based grounding, the MSR-based grounding model supports
incremental and continuous updates from observations and human
feedback. Assume temporarily that the MSR module receives a pair of
point cloud clusters corresponding to two objects, and the
prepositions of the spatial relations between the objects, e.g., from
QSR or humans. Our MSR module grounds each preposition using
histograms, also referred to as ``visual words'', which are created by
considering the point cloud data in a spherical coordinate system.
Specifically, each point is represented by its distance to a reference
point and two angles: $\theta\in [0\degree, 180\degree]$ and
$\varphi\in [-180\degree, 180\degree]$. On a robot, the coordinate
frame for grounding is defined with respect to the robot's coordinate
frame, its camera, and/or reference objects---information in one
coordinate frame can be transformed to other coordinate frames as
needed. Also, although processing the sensor input(s) can introduce
noise, the non-monotonic logical reasoning and incremental learning
abilities of our architecture support recovery from associated errors,
as described later.

%\begin{figure}[ht]
%  \centering
%  \includegraphics[width=.2\textwidth,height=4cm]{spherical_coord}
%  \caption{Spherical coordinates system}
%  \label{fig:spherical}
%\end{figure}  

\begin{figure}[tb]
  \centering
  \begin{subfigure}[b]{.44\textwidth}
    \includegraphics[width=\textwidth,height=4.7cm]{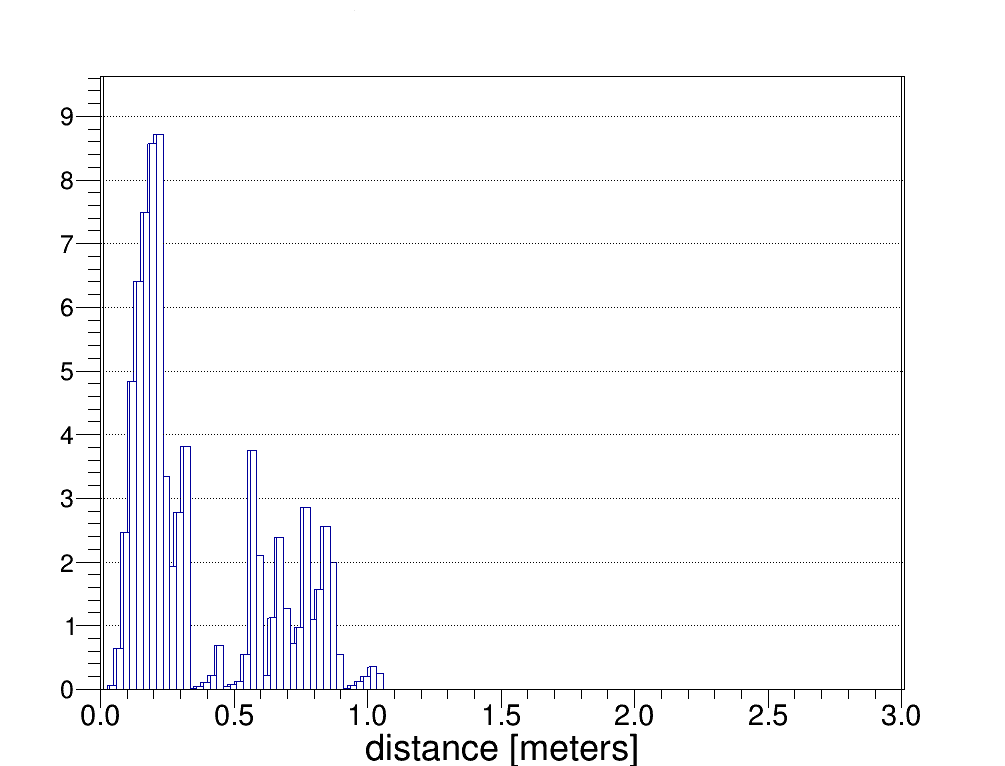}
    \caption{1D histogram for ``not-touching''.}
    \label{fig:hist_nontouch}
  \end{subfigure} 
  \hspace{4.5em}
  \begin{subfigure}[b]{.44\textwidth}
    \centering
    \includegraphics[width=\textwidth,height=4.7cm]{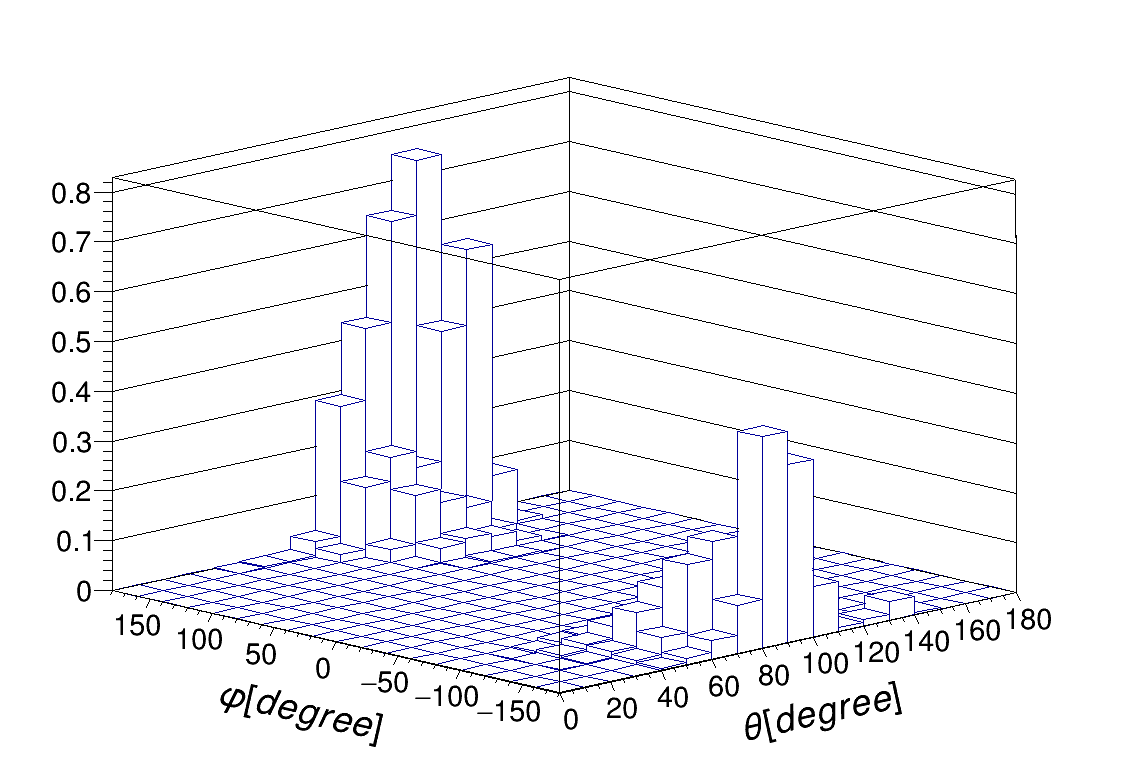}
    \caption{2D histogram for ``left''.}
    \label{fig:hist_left}
  \end{subfigure}
  \vspace{-1em}
  \caption{Illustrative examples of MSR-based grounding: (a) 1D
    histogram grounding distance-based preposition ``not-touching'';
    (b) 2D histogram grounding position-based preposition ``left''.}
  \vspace{-1em}
\end{figure}

In our MSR-based representation, each of the seven position-based
prepositions (\textit{in, left, right, front, behind, above, below})
are ground using 2D histograms of angles $\theta$ and $\varphi$,
whereas each of the three distance-based prepositions
(\textit{touching, not\textnormal{-}touching, far}) are ground using
1D histograms of the $10\%$ closest distances between points in pairs
of objects.  Figures~\ref{fig:hist_nontouch} and~\ref{fig:hist_left}
show one illustrative example (each) of a histogram grounding a
distance-based and position-based preposition. All histograms are
normalized to ensure that large objects with many points do not have
any undue influence on the grounding of relations.
 
Any learned MSR-based groundings are used for the subsequent new
scenes. For any given pair of point cloud clusters in a new scene, the
corresponding 2D and 1D histograms (i.e., visual words) are
constructed. The learned visual words that are most similar to the
extracted visual words are used to assign the distance-based and
position-based spatial relations between the corresponding scene
objects, e.g., ``$object_1$ is below $object_2$ and not touching it''.
These inferred spatial relations are automatically translated to logic
statements that are added to the ASP program, e.g.,
$obj\_relation(below,~obj_1,~obj_2)$.  Since axioms in the ASP program
are applied recursively for inference, each point cloud cluster only
needs to be considered once.

The similarity between visual words is computed using the standard
\emph{intersection} measure for 1D (distance) histograms. For
computing the similarity between 2D (position) histograms, we use the
$\chi^{2}$ measure, i.e., for two histograms $H$ and $G$:
%\vspace{-1em}
\begin{align}
  D_{\chi^{2}}(H,G) = \sum\limits_{i} \dfrac{\vert
    h_{i}-g_{i}\vert^{2}}{2(h_{i}+g_{i})}
  \label{eqn:chi-squared}
\end{align}
where $h_{i}$ and $g_{i}$ are bins in $H$ and $G$ respectively.
Smaller values of this measure denote greater similarity.  We only use
this measure for 2D histograms because the boundaries between
position-based relations are more difficult to define than those
between distance-based relations. Once the spatial relations between a
pair of point cloud clusters have been determined in a new scene, the
learned visual words are updated using a standard normalized histogram
merging approach, i.e., the \emph{MSR-based grounding is updated
  continuously}. Next, we consider the use of human feedback when it
is available.

%%%%-----------------------------------------------------------------------------
\subsubsection{Combined QSR-MSR Model}
\label{sec:arch-ground-discuss}
Recall that we are focusing on applications where many training
examples and human supervision are not readily available. However, we
do want to use the rich information encoded in human feedback when it
is available. While the QSR-based grounding remains unchanged, the
MSR-based grounding changes as new scenes are processed. Since the
QSR-based and MSR-based groundings may disagree on the relation
between some pairs of objects, the control node initially assigns high
(low) confidence to the QSR-based (MSR-based) grounding. The relative
confidence in each grounding is then updated based on the number of
times the output from the grounding matches human input---the more
reliable grounding is then used for processing the subsequent scenes.
Note that this approach assumes that human input of spatial relations
between point cloud clusters is accurate most of the time, i.e., each
human participant providing feedback is expected to interpret spatial
relations correctly. Incorrect human annotation can affect the
confidence in a grounding and the subsequent grounding of spatial
relations between objects, but our approach ensures that this only
happens if the number of such incorrect annotations is comparable to
the number of correct annotations.

Object shapes and sizes may also influence spatial relations depending
on the viewpoint. However, since the MSR-based grounding is based on
histograms of relative distances and angles, it can be used to infer
spatial relations over a range of viewpoints. Also, the architecture
has two mechanisms to limit and recover from errors. First, if the
QSR-based grounding is applicable, e.g., viewpoint has not changed
substantially from the initial view, the system can use it to obtain
an initial estimate of spatial relations and incrementally acquire the
MSR-based grounding. Second, if the QSR-based grounding is not
applicable, it is still possible to acquire an MSR-based grounding
from a small number of images and limited human input, and to use it
for subsequent inference.
 
There are some caveats related to the proposed approach.  First, the
QSR-based grounding is assumed to be reasonably accurate initially. If
this assumption does not hold and no human input is available, an
inaccurate MSR-based grounding may be acquired, resulting in incorrect
estimates of spatial relations. Also, it is possible to use an
accurate MSR-based grounding or human input to revise the QSR-based
grounding; we do not pursue that option in this paper in order to
simplify the process of understanding the two different groundings.
Second, human feedback improves the specialized MSR-based grounding
and overall accuracy, but it is not essential for estimating spatial
relations.  Third, the encoded prepositions (with learned grounding)
are translated to logic statements (i.e., observation literals) in an
ASP program. These observations and the commonsense knowledge encoded
in the ASP program limit possible relations between scene objects and
help infer composite relations (e.g., \textit{on, close to, next to}
etc). For instance, the spatial relation $on$ may be defined by the
following axiom:
\begin{align}
  obj\_relation(on, O_1, O_2)\ \leftarrow &\ obj\_relation(above, O_1,
  O_2),~obj\_relation(touching, O_1, O_2)
  \label{eqn:on}
\end{align}
which states that if object $O_1$ is above $O_2$ and touching it, then
$O_1$ is on $O_2$; the syntax and semantics of axioms are described in
the next section. However, all axioms need not be defined in advance;
our overall architecture supports reasoning with incomplete knowledge
and incremental learning of these axioms as described later in this
paper. Finally, we currently assume that each pair of objects is
related through one position-based and one distance-based spatial
relation, but not all the prepositions are (or need to be) mutually
exclusive.

%%%%%%%%%%%%%%%%%%%%%%%%%%%%%%%%%%%%%%%%%%%%%%%%%%%%%%%%%%%%%%%%%%%%%%%%%%%%%%%%%%%%
\subsection{Knowledge Representation and Reasoning}
\label{sec:arch-kr}
This section describes our approach for representing and reasoning
with incomplete domain knowledge. First, Section~\ref{sec:arch-kr-al}
introduces the \emph{action language} used in our architecture. Next,
Section~\ref{sec:arch-kr-ASP} describes the use of the action language
to represent a dynamic domain, and the translation of this domain
representation to an ASP program for non-monotonic logical inference.

%%%%-----------------------------------------------------------------------------
\subsubsection{Action Language}
\label{sec:arch-kr-al}
Action languages are formal models of part of a natural language used
for describing transition diagrams of dynamic domains. Our
architecture uses the action language
$\mathcal{AL}$~\cite{gelfond:ANCL13}, which has a sorted signature
with three types of sorts: \emph{statics}, which are domain attributes
whose values do not change over time; \emph{fluents}, which are domain
attributes that can be changed; and \emph{actions}. Fluents can be
\emph{inertial}, which can be directly modified by actions and obey
the laws of inertia, or \emph{defined}, which are not directly changed
by actions and do not obey inertia laws. A domain literal is a domain
attribute $\emph{p}$ or its negation $\emph{$\lnot$p}$. $\mathcal{AL}$
allows three types of statements:
%\begin{enumerate}
%\item:
\begin{align*}
  \emph{l} ~~\textbf{if} ~~ \emph{p}_{0},\ldots,\emph{p}_{m}~~~~~~~~&\textrm{State Constraints} \\
%\item Causal Laws:
  \emph{a} ~~\textbf{causes}\ ~~ \emph{l}_{in}~~ \textbf{if}~~
  \emph{p}_{0},\ldots,\emph{p}_{m}~~~~~~~~&\textrm{Causal Laws}\\
%\item :
%\begin{align}
  \textbf{impossible}~~ \emph{a}_{0},\ldots,\emph{a}_{k}
  ~~\textbf{if} ~~ \emph{p}_{0},\ldots,\emph{p}_{m}~~~~~~~~&\textrm{Executability Conditions}
\end{align*}
%\end{enumerate}
where $\emph{a}$ is an action, $\emph{l}$ is a literal,
$\emph{l}_{in}$ is an inertial literal, and $\emph{p}_{0}, \ldots,
\emph{p}_{m}$ are domain literals. Our architecture uses
$\mathcal{AL}$ to describe the transition diagram of any given domain,
as described below.

%%%%-----------------------------------------------------------------------------
\subsubsection{Knowledge Representation and Reasoning in ASP}
\label{sec:arch-kr-ASP}
A domain's description in $\mathcal{AL}$ comprises a \emph{system
  description} $\mathcal{D}$ and a \emph{history} $\mathcal{H}$.
$\mathcal{D}$ comprises a \emph{sorted signature} $\Sigma$ and axioms.
$\Sigma$ includes the \emph{basic sorts} arranged hierarchically,
\emph{domain attributes} (i.e., statics and fluents), and
\emph{actions}. In the RA domain, sorts include $object$, $robot$,
$entity$, $size$, $relation$, and $surface$, and the sort $step$ for
temporal reasoning, with $object$ and $robot$ being subsorts of
$entity$. Statics include some object attributes such as
$obj\_size(object, size)$ and $obj\_surface(obj, surface)$. Fluents of
the form $obj\_relation(relation, object, object)$ model relations
between objects in terms of their arguments' sorts, e.g.,
$obj\_relation(above, A, B)$ implies object $A$ is \emph{above} object
$B$---the last argument in these relations is the reference object for
the spatial relation under consideration. Fluents also describe other
aspects of the domain such as:
%\vspace{-0.5em}
%\begin{align}
$in\_hand(robot, object)$ and $stable(object)$,
%\end{align}
which describe whether a robot is holding a particular object, and
whether a particular object is stable (respectively). Actions of the
domain include $pickup(robot, object)$ and $putdown(robot, object,
location)$. A \emph{state} of the domain is then a collection of
ground literals, i.e., statics, fluents, actions, and relations with
values assigned to their arguments.

The axioms of $\mathcal{D}$ are defined in terms of the signature
$\Sigma$ and govern domain dynamics. These axioms include a
distributed representation of the constraints related to domain
actions, i.e., causal laws and executability conditions that define
the preconditions and effects of actions, and constraints related to
states, i.e., state constraints. The axioms of the RA domain include
statements such as:
\begin{subequations}
  \label{eqn:axiom-al}
  \begin{align}
    &pickup(robot, object)~~\textbf{causes}~~in\_hand(robot, object) \\
    &obj\_relation(below, B, A)~~\textbf{if}~~obj\_relation(above, A, B)\\
    &obj\_relation(behind, B, A)~~\textbf{if}~~obj\_relation(infront, A, B)\\
    &\textbf{impossible}~~ pickup(robot, object)~~ \textbf{if}~~in\_hand(robot, object)\\
    &\textbf{impossible}~~ putdown(robot, object, location)~~ \textbf{if}~~not~in\_hand(robot, object)
  \end{align}
\end{subequations}    
where Statement~\ref{eqn:axiom-al}(a) is a causal law which states
that if the robot executes the $pickup$ action on an object, it ends
up holding the object. Statements~\ref{eqn:axiom-al}(b-c) describe
state constraints regarding some spatial relations between two
objects. Statement~\ref{eqn:axiom-al}(d) describes an executability
condition which indicates that a robot cannot pick up an object that
it is already holding. Statement~\ref{eqn:axiom-al}(e) describes an
executability condition that uses default negation (i.e., $not$) in
the body of the axiom. This encodes a stronger constraint than the use
of classical negation (i.e., $\lnot$). This statement implies that it
is impossible for a robot to put a particular object down in a
particular location if it does not know whether the object is in its
hand or not, and not just when it is sure that it is not in its hand.

A history $\mathcal{H}$ of a dynamic domain typically includes records
of observations of fluents at particular time steps, i.e.,
$obs(fluent, boolean, step)$, and actions actually executed by the
robot at particular time steps, i.e., $hpd(action, step)$. In robotics
domains, it is common to have some default knowledge that holds in all
but a few exceptional circumstances. In other work from our research
group, we expanded the notion of history to include default statements
describing the values of fluents in the initial
state~\cite{mohan:JAIR19}.  For example, we may encode in the RA
domain that ``structures with four or more blocks are usually
unstable''.

To reason with the encoded domain knowledge, we construct the
CR-Prolog/ASP program $\Pi(\mathcal{D}, \mathcal{H})$ from the system
description $\mathcal{D}$ in $\mathcal{AL}$ and the history
$\mathcal{H}$. ASP is a declarative language that can represent
recursive definitions, defaults, causal relations, special forms of
self-reference, and language constructs that occur frequently in
non-mathematical domains, and are difficult to express in classical
logic formalisms. ASP is based on stable model
semantics~\cite{gelfond:aibook14} and supports concepts such as
\emph{default negation} (negation by failure) and \emph{epistemic
  disjunction}, e.g., unlike ``\stt{$\lnot$a}'', which implies that
``\emph{a is believed to be false}'', ``\stt{not a}'' only implies
``\emph{a is not believed to be true}''. Each literal can be true,
false, or unknown and the \emph{robot only believes that which it is
  forced to believe}. Unlike classical first order logic, ASP supports
non-monotonic logical reasoning, i.e., adding a statement can reduce
the set of inferred consequences, aiding in the recovery from errors
and situations in which observations do not match expectations due to
reasoning with incomplete knowledge; this is an essential capability
in robotics. ASP and other knowledge-based reasoning paradigms are
often criticized for requiring considerable prior knowledge, and for
being unwieldy in large, complex domains.  However, modern ASP solvers
support reasoning with incomplete knowledge and efficient reasoning in
large knowledge bases. These solvers and related reasoning systems are
used by an international research community in
robotics~\cite{erdem:KI18} and other applications~\cite{erdem:AIM16}.

A custom-built script is used to automate the translation of
$\mathcal{D}$ and $\mathcal{H}$ to $\Pi(\mathcal{D}, \mathcal{H})$.
The program $\Pi$ includes the signature and axioms of $\mathcal{D}$,
inertia axioms, reality checks (to ensure observations are consistent
with current beliefs), closed world assumptions for defined fluents
and actions, and observations, actions, and defaults from
$\mathcal{H}$. For instance, Statements~\ref{eqn:axiom-al}(a-e) of
$\mathcal{AL}$ are translated to:
\begin{subequations}
  \label{eqn:axiom-meta}
  \begin{align}
    holds(in\_hand(robot, object), I+1) &~\leftarrow~
    occurs(pickup(robot, object), I)\\
    holds(obj\_relation(above, A, B), I) &~\leftarrow~
    holds(obj\_relation(below, B, A), I) \\
    holds(obj\_relation(infront, A, B), I) &~\leftarrow~
    holds(obj\_relation(behind, B, A), I) \\
    \lnot occurs(pickup(robot, object), I) &~\leftarrow~
    holds(in\_hand(robot, object), I)\\
    \lnot occurs(putdown(robot, object, location), I) &~\leftarrow~
    not~holds(in\_hand(robot, object), I)
  \end{align}
\end{subequations} 
where the predicate $holds(fluent, step)$ implies that a particular
fluent holds true at a particular timestep, and the predicate
$occurs(action, step)$ implies that a particular action is supposed to
be executed at a particular time step in a plan.
% Other terminology includes: variable and object constants are
% \emph{terms}; terms with no symbols or variables are \emph{ground}.
% A predicate of terms is an \emph{atom}; it is \emph{ground} if all
% its terms are ground. An atom or its negation is a \emph{literal}:
% ground atoms and their negations are \emph{ground literals}.
In the context of the scene understanding tasks in the RA domain, the
program encodes prior knowledge about stability using axioms such as:
\begin{align}
  \label{eqn:axiom-unstabdef}
  \nonumber
  \neg holds(stable(A), I) \leftarrow& ~holds(obj\_relation(above, A,
  B), I),~size(A, large),\\
  &~size(B, small), ~not~ holds(stable(A), I)
\end{align}
which states that larger objects on smaller objects are unstable
unless there is evidence to the contrary. The spatial relations
extracted from RGB-D images are also automatically converted to
statements in ASP program that describe the current domain state.
Please see~\cite{gelfond:aibook14} for further examples of translating
a system description in $\mathcal{AL}$ to ASP. An illustrative example
of a complete program for the RA domain is in our open source
repository~\cite{code-results}.
%\footnote{\url{https://github.com/tmot987/RSS2019}}.

Once $\Pi(\mathcal{D}, \mathcal{H})$ is constructed, all reasoning,
i.e., planning, diagnostics, and inference can be reduced to computing
\emph{answer sets} of $\Pi$. These answer sets of $\Pi$ represent
beliefs of the robot associated with $\Pi$; this would include
inferred beliefs, plans, and results of diagnostics, as appropriate.
In the RA domain, an answer set could include beliefs about the
occlusion of individual objects and the stability of object
structures, and a plan to reduce clutter. To compute the answer set(s)
of any given ASP program, we use the SPARC
system~\cite{balai:lpnmr13}, which is based on a SAT solver. The
computation of answer set for planning or diagnostics also requires us
to introduce some helper axioms, which would include a goal definition
and the following for planning:
\begin{subequations}
    \label{eqn:plan-helper}
    \begin{align}
    &success \leftarrow goal(I) \\
    &\leftarrow not~~success \\ 
    &occurs(A,I) ~~|~~ \lnot occurs(A,I) \leftarrow not~~goal(I)\\
    &\leftarrow occurs(A_1,I), ~occurs(A_2,I), ~A_1 \neq A_2 \\
    & something\_happened(I) \leftarrow occurs(A,I) \\
    &\leftarrow goal(I), ~goal(I-1), ~J < I, ~not~~something\_happened(J)
    \end{align}
\end{subequations}
which force the robot to search for actions until the goal is achieved
and prevent the robot from executing multiple actions concurrently.
Some axioms are also introduced such that unexpected observations
result in an inconsistency that the robot resolves using
\emph{Consistency Restoring} (CR) rules~\cite{balduccini:aaaisymp03}.
We do not include all these axioms here but the ASP program for the RA
domain is in our code repository~\cite{code-results}.

Since the robot only believes that which it is forced to believe, the
inability to compute an answer set indicates an unresolved
inconsistency that is considered to be due to incomplete knowledge or
an error in the encoding that needs to be probed further. In the
context of the scene understanding tasks under consideration, the
robot would either be unable to make a decision regarding occlusion
and stability, or provide an incorrect inference (when ground truth is
available). This situation is addressed in our architecture using the
attention mechanism and deep networks as described below.

%%%%%%%%%%%%%%%%%%%%%%%%%%%%%%%%%%%%%%%%%%%%%%%%%%%%%%%%%%%%%%%%%%%%%%%%%%%%%%%%%%%%
\subsection{Attention Mechanism and Deep Learning}
\label{sec:arch-roicnn}
The attention mechanism module is used when ASP-based reasoning is
unable to assign (occlusion and stability) labels to objects in the
input image, or when it assigns an incorrect label (for training
data). In each such image, this module automatically directs attention
to regions of interest (ROIs) that contain information relevant to the
task at hand.  To do so, it identifies each axiom in the ASP program
whose head corresponds to a relation relevant to the task at hand.
The relations in the body of each such selected axiom are then used to
identify image ROIs to be processed further; the remaining image
regions are unlikely to provide relevant information and are not
analyzed further. Note that our formulation of ``attention'' draws on
early work in computer vision, AI, and psychology; it is not based on
how this concept has been modeled in the deep learning literature.

%\emph{The key contribution is that the process of identifying ROIs using current knowledge is fully automated}.

\begin{figure}[tb]
  \begin{center}
    \begin{subfigure}{0.45\textwidth}
      \includegraphics[width=\textwidth]{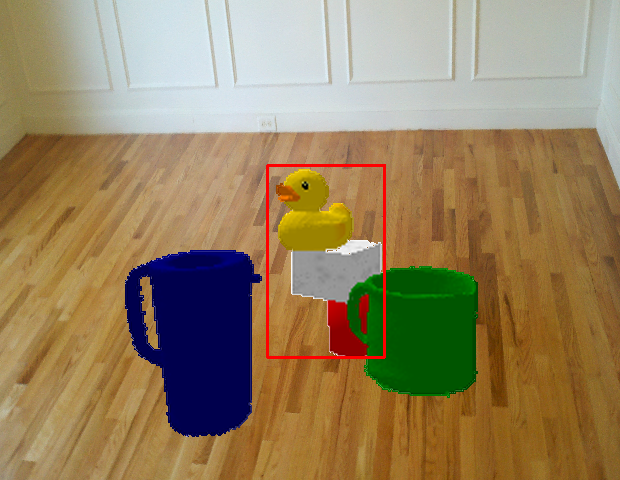}
      \caption{}%Example image of scene.}
      \label{fig:att1}
    \end{subfigure}
    \hspace{0.1in}
    \begin{subfigure}{0.45\textwidth}
      \includegraphics[width=\textwidth]{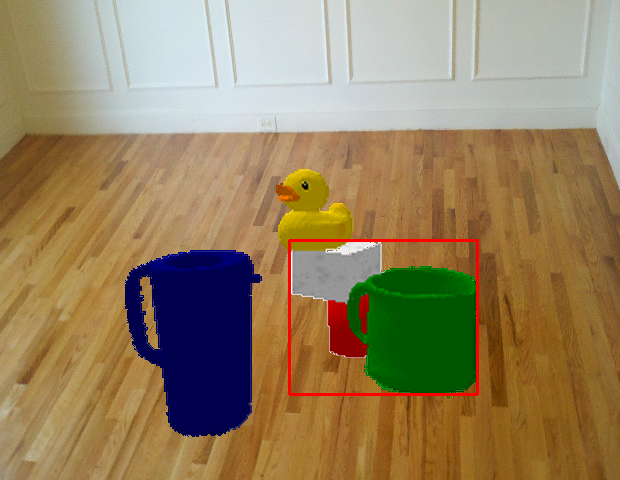}
      \caption{}%Point clouds of scene.}
      \label{fig:att2}
    \end{subfigure}
  \end{center}
  \vspace{-1.5em}
  \caption{Examples of ROIs automatically identified by the attention mechanism for
    further analysis in the context of estimating: (a) stability of
    object structures; and (b) occlusion of objects.}
  \label{fig:attention}
  \vspace{-1em}
\end{figure}

As an example, consider the task of determining the stability of
object structures in the image in Figure~\ref{fig:att1}.  Axioms that
define conditions under which a particular object is considered to be
stable, and those that define conditions under which an object is
considered to be unstable, are relevant to this task. These axioms
would have $stable(A)$ or $\lnot stable(A)$ in the head of the axiom,
e.g., Statement~\ref{eqn:axiom-learn}(a) and
Statement~\ref{eqn:axiom-unstab} respectively in
Section~\ref{sec:arch-dtmerge} below. The head of any such axiom holds
true in any state in which all the relations in the body of the axioms
are satisfied. In the case of Statement~\ref{eqn:axiom-learn}(a) and
Statement~\ref{eqn:axiom-unstab}, the body of the axiom contains the
spatial relation $above$, leading the attention mechanism to consider
the stack comprising the duck, the red can, and the white cube, as
indicated by the red rectangle in Figure~\ref{fig:att1}, because they
satisfy the relevant relation. While this ROI is analyzed further,
other image regions and objects (e.g., mug, pitcher) are disregarded.
Note that \emph{the process of selecting axioms in the knowledge base
  relevant to any given task, and extracting image ROIs satisfying the
  body of these axioms, is fully automated}; there is no manual
intervention.

As another example, consider the task of identifying occluded objects
in Figure~\ref{fig:att2}.  Statement~\ref{eqn:axiom-learn}(b) in
Section~\ref{sec:arch-dtmerge}, which defines conditions under which
an object is not considered to be occluded (as indicated by the
axiom's head), is relevant to the task. This axiom's body indicates
that the relation $behind$ is relevant for decisions about occlusion
of objects, and the attention mechanism will only consider pairs of
objects in Figure~\ref{fig:att2} that satisfy this relation. The red
rectangle in Figure~\ref{fig:att2} indicates the relevant image region
comprising the mug, the red can, and the white cube. This region is
analyzed further, whereas the other image regions are disregarded. As
before, this selection of the ROI is achieved without any manual
supervision.

Once the attention mechanism identifies image ROIs enveloping objects
that could not be assigned stability and occlusion labels using
ASP-based reasoning, pixels of each such ROI are considered to provide
information that is relevant to the estimation tasks but is not
captured by the existing knowledge base (i.e., ASP program). The
pixels in these ROIs and the target labels to be assigned to objects
(and structures) in the ROIs are provided as inputs to a CNN.  The CNN
learns the mapping between the image pixels and target labels, and
then assigns these labels to ROIs in previously unseen test images
that ASP-based reasoning is unable to process.

A CNN has many parameters based on size, number of layers, activation
functions, and the connections within and between layers, but the
basic building blocks are convolutional, pooling, and fully-connected
layers. The convolutional and pooling layers are used in the initial
or intermediate stages of the network, whereas the fully-connected
layer is typically one of the final layers. In a convolutional layer,
a filter (or kernel) is convolved with the original input (to the
network) or the output of the previous layer. One or more
convolutional layers are usually followed by a pooling layer. Common
pooling strategies such as max-pooling and average-pooling are used to
reduce the dimensions of the input data, limit the number of
parameters, and control overfitting. The fully-connected layers are
equivalent to feed-forward neural networks in which all neurons
between adjacent layers are connected---they often provide the target
outputs. In the context of images, convolutional layers extract useful
attributes to model the mapping from inputs to outputs, e.g., the
initial layers may extract lines and arcs, whereas the subsequent
layers may compose more complex geometric shapes. While estimating the
stability of object configurations, the CNN's layers may implicitly
represent attributes such as whether: (i) a tower of blocks is
aligned; (ii) an object with an uneven surface is under another
object; or (iii) a tower has a small base.

\begin{figure}[tb]
 \begin{center}
  \includegraphics[width=0.7\textwidth]{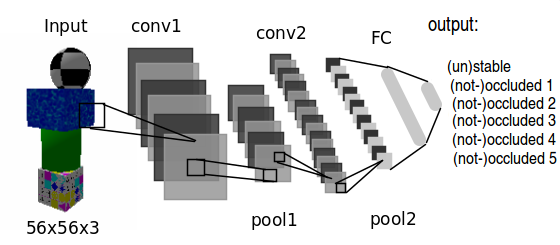}
  \vspace{-1em}
  \caption{Lenet architecture.}
  \label{fig:lenet_arch}
  \vspace{-2em}
  \end{center}
\end{figure}

In this paper, we adapt and use two established CNN architectures: (i)
Lenet~\cite{lecun1998}, initially proposed for recognizing
hand-written digits; and (ii) Alexnet~\cite{krizhevsky2012}, which has
been used widely since it provided very good results on the Imagenet
benchmark dataset. The Lenet has two convolutional layers, each one
followed by a max-pooling layer and an activation layer. Two fully
connected layers are placed at the end.  Unlike the $28\times 28$
gray-scale input images and the ten-class softmax output layer used in
the original implementation, we consider $56\times 56$ RGB images as
the input. Note that each such input to the network corresponds to an
image ROI under consideration. The input vector size was chosen
experimentally using validation sets and ROIs were scaled
appropriately; the minimal improvement in performance provided by
longer input vectors did not justify the significant increase in
computational effort. The network's outputs estimate the occlusion of
each object and the stability of object structure(s) in the ROI under
consideration. As described later in Section~\ref{sec:expres-setup},
we consider ROIs with up to five objects, and we use the sigmoid
activation function.  Figure~\ref{fig:lenet_arch} is a pictorial
representation of this network. The Alexnet architecture, on the other
hand, contains five convolutional layers, each followed by max-pooling
and activation layers, along with three fully connected layers at the
end.  In our experiments, each input vector is a $227\times 227$ RGB
image. The size of the output vector and the activation function are
the same as those for the Lenet architecture.

With both CNN architectures, we used the Adam
optimizer~\cite{kingma2014} in TensorFlow~\cite{abadi2016} with a
learning rate of $0.0002$ and $0.0001$ for Lenet and Alexnet
respectively; the initial weights were initialized randomly. The
number of training iterations varied depending on the network and the
number of training examples. For example, the Lenet network using
$100$ ($5000$) image samples was trained for $10000$ ($40000$)
iterations, whereas the Alexnet with $100$ ($5000$) training samples
was trained for $8000$ ($20000$) iterations. The learning rate and
number of iterations were chosen experimentally using validation sets.
The number of epochs was chosen as the stopping criteria, instead of
the training error, in order to allow the comparison between networks
learned with and without the attention mechanism. The code for
training the deep networks is in our open source
repository~\cite{code-results}. Note that other, potentially more
sophisticated, deep network models could be used in our overall
architecture (instead of Lenet or Alexnet), but this is beyond the
scope of this work. Also, the chosen CNN architectures are sufficient
for the learning task when it is informed by reasoning with
commonsense knowledge.

Recall that a CNN is only trained on ROIs from images for which
ASP-based reasoning provides an incorrect outcome or is unable to
provide an outcome. We consider any such trained CNN to represent
previously unknown knowledge not encoded, or encoded incorrectly, in
the ASP program. In other words, the observed incorrect outcome or
lack of any outcome is considered to be a consequence of reasoning
with incomplete or incorrect knowledge, which (in turn) can be because
the knowledge was incomplete or incorrect when it was encoded
initially, or because of changes in the domain over time. The next
component of our architecture supports the incremental acquisition of
previously unknown state constraints from the ROIs (used to train the
deep networks), and the merging of this information with the existing
knowledge. This process also (indirectly) helps understand the
behavior of the trained deep networks.

%%%%%%%%%%%%%%%%%%%%%%%%%%%%%%%%%%%%%%%%%%%%%%%%%%%%%%%%%%%%%%%%%%%%%%%%%%%%%%%%%%%%
\subsection{Decision Tree Induction and Axiom Merging}
\label{sec:arch-dtmerge}
In our architecture, previously unknown domain knowledge is learned
incrementally using decision tree induction, and a heuristic approach
inspired by human forgetting merges the learned knowledge with the
existing knowledge for subsequent reasoning. As stated in
Section~\ref{sec:introduction}, we illustrate this capability in this
paper by learning axioms that represent previously unknown state
constraints. In other work, we have demonstrated the use of the
inductive learning approach, without the heuristic merging strategy,
to learn other kinds of axioms~\cite{mota:aaaisymp20}.

We adapt the well-known ID3 algorithm~\cite{quinlan1986} to construct
the desired decision trees, using entropy minimization as the
criterion to select the attribute to split the nodes. Building on the
underlying distributed, relational representation of axioms, we learn
separate decision trees for each estimation task, e.g., in the RA
domain, we learn separate decision trees for stability estimation and
occlusion estimation. The difference in the construction of these
decision trees is in the use of the relational descriptions based on
prior knowledge and the observations as the attributes. Specifically,
relational domain attributes are extracted automatically from the
image ROIs used for training the deep networks; recall that these
attributes include the spatial relations between pairs of objects and
the attributes of the objects in each such ROI. This relational
information and the corresponding occlusion and stability labels are
used as the labeled training examples to build the decisions trees for
the estimation tasks under consideration. For the illustrative example
domain considered in this paper, each tree's nodes encode splits based
on the domain attributes, and the labels of the leaf nodes are
\emph{stable}, \emph{unstable}, \emph{occluded}, and \emph{not
  occluded}.

\begin{figure*}[tb]
  \begin{center}
    \includegraphics[width=1\linewidth]{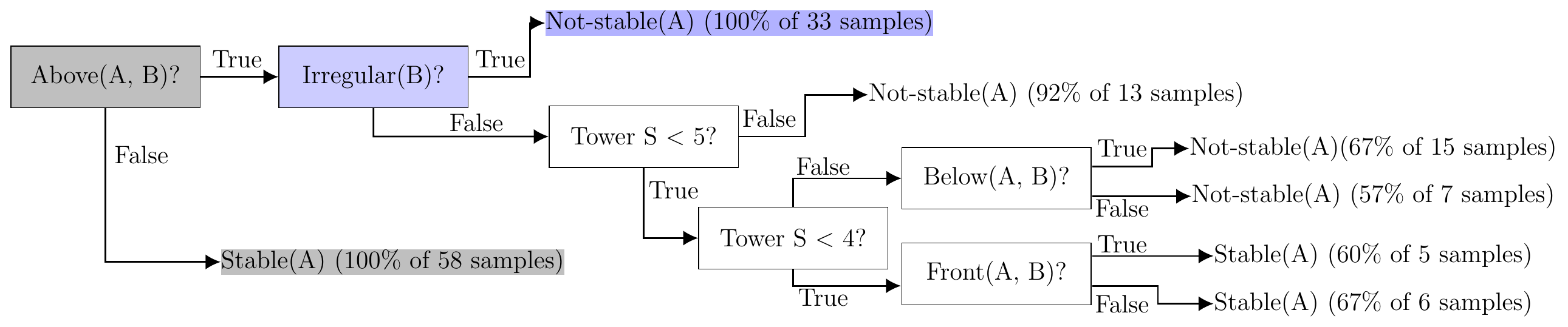}
    \caption{Example of a decision tree constructed for stability
      estimation using some labeled examples. Highlighted branches are
      used to construct previously unknown axioms.}
    \label{fig:decisiontree-example}
  \end{center}
  \vspace{-1em}
\end{figure*}

\begin{figure*}[tb]
  \begin{center}
    \includegraphics[width=0.95\linewidth]{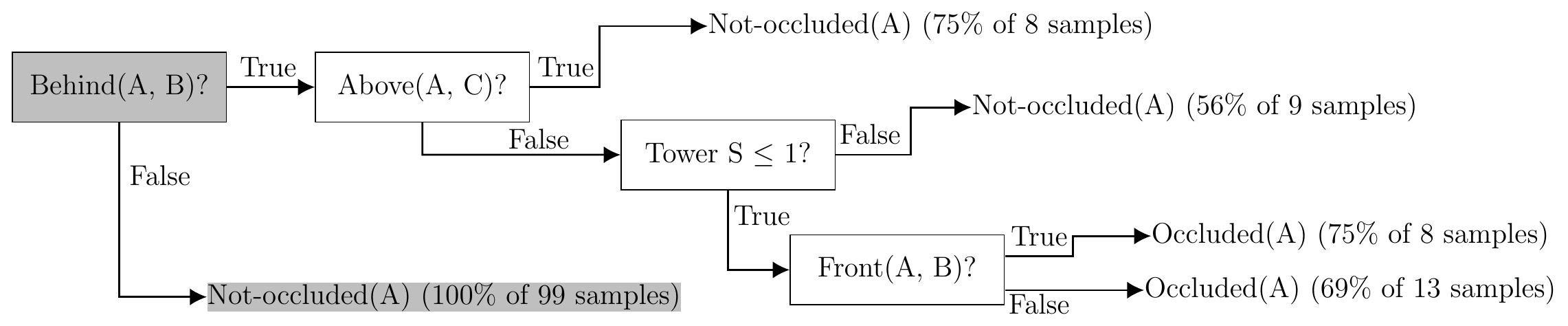}
    \caption{Example of a decision tree constructed for occlusion
      estimation using some labeled examples. Highlighted branch is
      used to construct previously unknown axiom.}
    \label{fig:decisiontree-example2}
  \end{center}
  \vspace{-1.5em}
\end{figure*}

Once the decision trees are constructed, our approach automatically
extracts candidate axioms from the trees. Consider, for example, the
branches highlighted in gray in Figures~\ref{fig:decisiontree-example}
and~\ref{fig:decisiontree-example2}, which show part of the decision
trees learned in the RA domain.  These branches can be translated to
the axioms:
\begin{subequations}
  \label{eqn:axiom-learn}
  \begin{align}
    &holds(stable(A), I) \leftarrow \ not~ holds(obj\_relation(above, A, B), I) \\
    &\neg holds(occluded(A), I) \leftarrow \ not~ holds(obj\_relation(behind, A, B), I)
  \end{align}
\end{subequations}
where Statement~\ref{eqn:axiom-learn}(a) implies that any object that
is not known to be above any other object is considered to be stable,
whereas Statement~\ref{eqn:axiom-learn}(b) says that an object is not
occluded if it is not known to be behind any other object. Note that
this translation from the decision tree to axioms uses default
negation to model the use of quantifiers with specific negated
literals in the body of axioms. As another example, the branch
highlighted in gray and blue in Figure~\ref{fig:decisiontree-example}
translates to the following axiom:
\begin{align}
  \label{eqn:axiom-unstab}
  \neg holds(stable(A), I) ~\leftarrow~ &holds(obj\_relation(above, A, B), I),\
  obj\_surface(B, irregular)
\end{align}
which states that an object is unstable if it is located above another
object with an irregular surface.

\begin{algorithm}[tbh]
\algsetup{linenosize=\normalsize}
\normalsize
\caption{\textbf{Learning and merging axioms}}
\label{alg:learning}
\DontPrintSemicolon
\SetKwInOut{Input}{Input}\SetKwInOut{Output}{Output}

\Input{Relational domain attributes from image ROIs; occlusion and
  stability labels of objects in ROIs; thresholds $th_1$ ($95\%$,
  purity threshold), $th_2$ ($5\%$, support threshold), $th_3$
  ($40\%$, tree support threshold), $th_4$ ($10\%$, axiom strength
  threshold); ensemble\_count ($100$).}

\Output{Learned axioms.}

%\Begin{
\While{true}{
  
  \If{labeled\_samples} {
    
    \For{$j = 1:ensemble\_count$}{
      \tcp{Split training samples for learning and validation}
      training\_set,\ validation\_set = random\_split(labeled\_samples)\;
      
      \tcp{Decision tree induction}
      learned\_tree = tree\_induction(training\_set)\;
      
      \tcp{Create candidate axioms}
      candidate\_axioms = select(learned\_tree, $th_1$, $th_2$)\;
      
      \tcp{Validate axioms}
      validated\_axioms = validate(candidate\_axioms, validation\_set, $th_2$)\;
      % \For{$sample\ in\ validation\_set$}{
      %   branch = find\_branch(sample,\ branches)\;
      %   nodes = split(branch)\;
      %   candidates.append(all\_combinations(nodes))\;
      % }
      
      % \tcp{Removing bad candidates}
      % filtered\_candidates = Filter(candidates)\;
      % \tcp{Counting each tree votes}
      % tree\_vote $+=$ count(filtered\_candidates)\;
    }
    \tcp{Choose validated axioms with sufficient support}
    axioms $=$ select(validated\_axioms, $th_3$)\;
    
    \tcp{Add validated axioms and merge similar axioms}
    add\_merge(axioms)\;
  } 
  
  \tcp{Update strength of axioms}
  update\_strength(axioms)\;

  \tcp{Remove axioms with low strength}
  remove(axioms, $th_4$)\;

  % Pool = Memory.group(similar\_Axioms)\;
  % \tcp{keeping only the best axiom from each pool using 10 labeled scenes}
  % Pool.select(best\_Axiom, 10 samples)\;
}
\end{algorithm}

Algorithm~\ref{alg:learning} describes the steps for automatically
constructing the decisions trees, extracting and validating candidate
axioms, and merging the valid new axioms with the existing knowledge.
The algorithm first checks for suitable labeled training examples,
i.e., image ROIs for which the stability and occlusion labels could
not be determined correctly using ASP-based reasoning; these are used
to induce new state constraints (lines 2-11). Specifically, a training
set is created by randomly selecting $50\%$ of the labeled examples,
with the remaining examples making up the validation set (line 4). As
stated earlier, the construction of the tree is based on the ID3
algorithm~\cite{quinlan1986}; it considers the relational attributes
extracted from the image ROIs under consideration to split nodes, and
assigns the known labels to the leaves. During the construction of the
tree, the algorithm computes the potential change in entropy (i.e.,
information gain) that would occur if a split is introduced at a node
in the tree based on each attribute that has not yet been used (line
5); the attribute that is likely to provide the highest reduction in
entropy is selected to split the examples at a node.

Given any such tree, the branches of the tree (from root to leaves)
that satisfy certain minimum requirements are selected to construct
candidate axioms (line 6). These minimum requirements include
thresholds on purity of samples at any given leaf, and on the support
from the labeled examples, e.g., $\ge 95\%$ examples at the leaf
belong to a particular (correct) label, and a branch under
consideration has support from $\ge 5\%$ of the training samples. The
selected branches of the learned decision trees represent previously
unknown candidate constraints, and the thresholds are set to construct
such candidate axioms cautiously, i.e., small changes in the value of
these thresholds do not cause any significant change in the branches
of the tree selected to construct axioms. The values of these
thresholds can also be revised intentionally to achieve different
desired behavior.  For instance, to identify default constraints that
hold in all but a few exceptional circumstances (see
Section~\ref{sec:arch-kr-ASP}), we lower the threshold for selecting a
branch of a tree to construct candidate axioms (from $95\%$ to
$70\%$). As we will discuss later, lowering the thresholds results in
the discovery of additional axioms, but also introduces noise. In
addition, in this paper, learning of axioms containing default
negation literals is restricted to default constraints such as
Equation~\ref{eqn:axiom-unstabdef}. In other words, the default
negation is only associated with a literal in the body that is the
negation of the literal in the head. This restriction is imposed
primarily for computational efficiency because associating default
negation with all literals would significantly increase the search
space. Also, in any particular domain, the default negation is only
applicable to certain literals, and this knowledge can be used during
axiom learning. So, we demonstrate the capabilities of our approach
with this restriction.

Once the candidate axioms are constructed, each one is validated using
the other half of the labeled examples that have (so far) not been
seen by the algorithm (line 7). The validation process removes axioms
without a minimum level of support (e.g., $5\%$) from the labeled
examples. Since the number of labeled examples available for training
is often small, we reduce the effect of noise through a homogeneous
ensemble learning approach (lines 3-8), i.e., we repeat the learning
and validation steps a number of times (e.g., $100$) and only the
axioms identified in more than a minimum number of iterations (e.g.,
$40\%$, line 9) are retained. Adding all retained axioms can lead to
the ASP program including different versions of the same axiom over
time. For instance, two axioms may have identical heads with one
axiom's body containing all the literals of the other, or two ground
axioms may include sorts that are subsorts of a more general sort. To
address this issue the algorithm reasons with the existing knowledge
to identify the axioms to be added to the knowledge base
($add\_merge(axioms)$, line 10). First, axioms with the same head and
overlap in the body are grouped together. Each possible combination of
axioms from different groups (one from each group at a time) is then
encoded in an ASP program along with the axioms that do not belong to
any such group. The resulting program is used to classify ten labeled
scenes chosen randomly. Axioms in the program that results in the
highest accuracy are retained whereas the other axioms in each group
are discarded.

% \begin{enumerate}[label=\arabic*)]
% \item \textbf{Wrong}: with the same head as the correct version, but
%   inverted (negated when it shouldn't be, and vice-versa) literal in
%   its body;
% \item \textbf{Incomplete}: if at least one literal is missing in the
%   axiom body;
% \item \textbf{Over-specified}: if one or more extra literals are
%   present in the axiom body when compared with its ideal version.
% \end{enumerate} 

The axiom learning approach described so far is based on a small
number of labeled examples in a dynamic domain. The learned axioms may
be incorrect (e.g., incorrect negation in the head, or incorrect
literals in the body), incomplete (e.g., one or more missing literals
in the body), or over-specified (e.g., one or more irrelevant literals
in the body). Reasoning with these axioms can lead to sub-optimal or
incorrect behavior. To address this issue, we incorporated a heuristic
approach inspired by the human forgetting behavior~\cite{ellwart2019}.
This approach associates a ``strength'' values to each axiom. An
axiom's strength is revised over time based on a decay factor using an
exponential relation: $axiom\_relevance = e^{-\alpha.n}$, where
$\alpha$ represents the decay factor (initially $1$), and $n$ is the
number of time steps since the axiom was learned.  In each time step,
irrespective of whether any new axioms are learned, the strength of
all learned axioms are updated (line 12). If an axiom is reinforced,
i.e., learned again or used, its strength is elevated to the maximum
value (i.e., $1$) again, and its decay factor is divided by
$\sqrt[n]{2}$, a value chosen experimentally such that it varies
between 2 (for $n=1$) and 1 (for $n \rightarrow \infty$). Any axiom
whose strength value falls below a threshold (e.g., $0.1$) is removed
from further consideration (line 13).

% Overall, the algorithm detailed in Table \ref{alg:learning} describes
% for each learning cycle (for loop starting in line 1) how features and
% labels from scenes are transcribed into axioms (using decision tree
% induction), and the subsequent procedure for including these axioms in
% the memory as well as removing obsolete information.

%%%%%%%%%%%%%%%%%%%%%%%%%%%%%%%%%%%%%%%%%%%%%%%%%%%%%%%%%%%%%%%%%%%%%%%%%%%%%%%%%%%%
%%%%%%%%%%%%%%%%%%%%%%%%%%%%%%%%%%%%%%%%%%%%%%%%%%%%%%%%%%%%%%%%%%%%%%%%%%%%%%%%%%%%

\section{Experimental Setup and Results}
\label{sec:expres}
In this section, we first describe the experimental setup for
evaluating our architecture (Section~\ref{sec:expres-setup}). We also
specify the hypotheses to be evaluated for the: (a) incremental
grounding of spatial relations; and (b) the estimation of object
occlusion and the stability of object structures.
Section~\ref{sec:exec-trace} describes some execution traces and
Section~\ref{sec:expres-results} discusses the results of experimental
evaluation. 

As stated in Section~\ref{sec:introduction}, our focus is on exploring
the interplay between reasoning and learning for reliable and
efficient operation in any given dynamic domain with a limited number
of labeled training examples. Benchmark datasets that include a large
number of images from different domains and approaches that do not
support the desired coupling between representation, reasoning, and
learning are thus not used for experimental evaluation.  Instead, our
experimental evaluation uses a limited number of real-world and
simulated images from the domain under consideration, and considers
ablation studies and architectures based just on deep networks as
baselines for experimental evaluation.

%%%%%%%%%%%%%%%%%%%%%%%%%%%%%%%%%%%%%%%%%%%%%%%%%%%%%%%%%%%%%%%%%%%%%%%%%%%%%%%%%%%%
\vspace{-1em}
\subsection{Experimental Setup}
\vspace{-0.5em}
\label{sec:expres-setup}
We first describe the experimental set up, datasets, and the
hypotheses for experimental evaluation. We do so separately for the
incremental grounding of spatial relations
(Section~\ref{sec:expres-setup-grounding}) and the estimation of
object occlusion and stability
(Section~\ref{sec:expres-setup-estimate}).

%%%%-----------------------------------------------------------------------------
\subsubsection{Incremental Grounding}
\begin{figure}[tb]
  \begin{center}
    \begin{subfigure}{0.35\textwidth}
      \includegraphics[width=\textwidth]{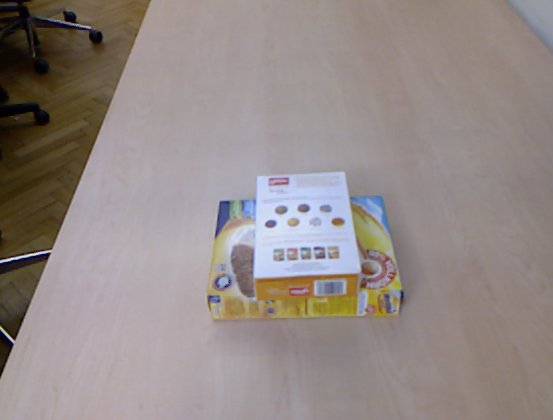}
      \caption{Scene with two objects.}
      \label{fig:tosd1}
    \end{subfigure}
    \hspace{0.1in}
    \begin{subfigure}{0.34\textwidth}
      \includegraphics[width=\textwidth]{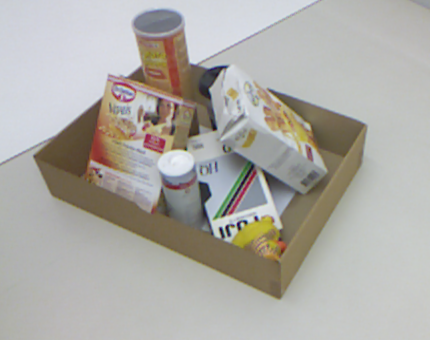}
      \caption{Complex scene with multiple objects.}
      \label{fig:tosd2}
    \end{subfigure}
  \end{center}
  \vspace{-1.5em}
  \caption{Examples of images from the TOSD dataset.}
  \label{fig:tosd-image}
  \vspace{-1em}
\end{figure}

\vspace{-0.5em}
\label{sec:expres-setup-grounding}
For evaluating the grounding of spatial relations, we used the Table
Object Scene Database
(TOSD)\footnote{https://repo.acin.tuwien.ac.at/tmp/permanent/TOSD.zip},
with $111$ scenes for training and $131$ scenes for testing. TOSD
contains scenes of real objects on a tabletop for evaluating
segmentation algorithms. Many scenes include complex object
configurations, e.g., Figure~\ref{fig:tosd2}, while some scenes have
only two objects, e.g., Figure~\ref{fig:tosd1}. We chose this dataset
because it provides a good combination of simple and complex scenes,
and has been used as a benchmark in other work on segmentation and
grounding of spatial relations. Since TOSD does not include spatial
relation labels, we manually labeled the relations between objects in
$200$ scenes. We then experimentally evaluated the following
hypothesis:
\begin{enumerate}
\item[\underline{\textbf{H1}}] The combination of manually-encoded QSR
  grounding and incrementally-learned MSR grounding performs better
  than each grounding used individually, and enables more effective
  use of human feedback.
\end{enumerate}
The performance measure for evaluating this hypothesis was the
accuracy of the labels assigned to the spatial relations between
objects in the scene under consideration. Note that the labels
assigned manually before the experimental evaluation provided the
ground truth, and the human feedback provided during the experiments
was in the form of labels for spatial relations between pairs of
objects.

%%%%-----------------------------------------------------------------------------
\subsubsection{Occlusion and Stability Estimation}
\label{sec:expres-setup-estimate}
For evaluating the assignment of occlusion and stability labels to
objects and object structures, we used a real-time physics engine
(Bullet physics library) to generate $6000$ images. The objects we
simulated included cylinders, spheres, cubes, a duck, and five objects
from the Yale-CMU-Berkeley dataset (apple, pitcher, mustard bottle,
mug, and cracker box)~\cite{Calli2015}. Objects were characterized by
different colors, textures, and shapes. We considered three different
arrangements of these objects:
\begin{itemize}
\item \textbf{Towers}: images containing $2-5$ objects stacked on top
  of each other;
\item \textbf{Spread}: images with five objects placed in different
  relative positions on a flat surface; and
\item \textbf{Intersection}: images with $2-4$ objects stacked on each
  other, with the rest ($1-3$ objects) on a flat surface.
\end{itemize}
The vertical alignment of stacked objects was randomized creating
either a stable or an unstable arrangement. The horizontal distance
between objects was also randomized, creating scenes with complex,
partial, or no occlusion. Lighting, orientation, camera distance,
camera orientation, and background, were also randomized.  An
additional $600$ labeled simulated scenes were also created and used
for evaluation, e.g., for the approach to update the strength of the
learned axioms. In addition, in the experimental trials summarized
below, the ASP program was initially missing three state constraints
(each) related to stability estimation and occlusion estimation.

A second dataset was derived from the dataset described above to
represent information about ROIs in the input images. Recall from
Section~\ref{sec:arch-roicnn} that the attention mechanism module
automatically extracts ROIs from input images that could not be
labeled using ASP-based reasoning, by identifying relevant axioms and
relations in the ASP program. Given the object arrangements described
above, any ROI in any given image can have up to five objects. The
second dataset automatically assigned labels to different possible
ROIs in the images, which is used as the ground truth. During
experimental evaluation, the application of the attention mechanism on
any given image identified objects of interest in this image. The
corresponding ROI from the second dataset was identified and the
information from this ROI was analyzed instead of information from the
entire image.

As baselines, the CNN architectures (Alexnet, Lenet) were trained and
evaluated on the first dataset without our reasoning and attention
mechanism modules. These were compared against the performance of our
architecture for different number of training samples. Recall that
occlusion is estimated for each object (i.e., maximum of five outputs
for a ROI) and stability is estimated for each object structure (i.e.,
one output for each structure/ROI).  The following hypotheses were
evaluated experimentally:
\begin{enumerate}
\item[\underline{\textbf{H2}}] Reasoning with commonsense domain
  knowledge and the attention mechanism to guide deep learning
  improves accuracy in comparison with the baselines.

\item[\underline{\textbf{H3}}] Reasoning with commonsense domain
  knowledge and the attention mechanism to guide deep learning reduces
  sample complexity and thus the computational effort in comparison with
  the baselines.
  
\item[\underline{\textbf{H4}}] Our architecture is able to
  incrementally learn previously unknown axioms, and using these
  axioms improves the accuracy of decision making in comparison with
  reasoning without the learned axioms.
  
\item[\underline{\textbf{H5}}] Our approach for revising the strength
  of axioms and merging similar axioms is able to identify and remove
  incorrect axioms.
\end{enumerate}
The main performance measures were the accuracy of the labels assigned
to objects and objects structures, and correctness of the axioms
learned and retained in the ASP program.
% As baselines for comparison, we trained the Lenet and Alexnet
% architectures without the commonsense reasoning and attention
% mechanism modules, i.e., on the RGB-D input images, and evaluated
% them on the test dataset.
Unless stated otherwise, \emph{all claims are statistically
  significant at $95\%$ significance level}. Before we discuss the
quantitative experimental results, we first describe the working of
our architecture using execution traces.

%%%%%%%%%%%%%%%%%%%%%%%%%%%%%%%%%%%%%%%%%%%%%%%%%%%%%%

\subsection{Execution Trace}
\label{sec:exec-trace}
The following two execution traces illustrate the reasoning, axiom learning, and axiom merging capabilities of our architecture. 

\begin{execexample}\label{exec-example1}[Planning and learning] \\
  {\rm Consider the scenario showed in Figure~\ref{fig:attention}. The
    robot is given the goal of placing the red can on top of the white
    block, i.e., \textit{holds(relation(on, red\_can, white\_block),
      I)}. The agent initially does not know that an object placed on
    another object with an irregular surface is unstable, i.e,
    Statement \ref{eqn:axiom-unstab}, and creates the following plan:
  \begin{enumerate}
  \item Pick up the duck in step 0;
  \item Put down the duck on the table in step 1;
  \item Pick up the white cube in step 2;
  \item Put down the white cube on the duck in step 3;
  \item Pick up the red can in step 4;
  \item Put down the red can on the white cube in step 5. 
  \end{enumerate}
  Since the robot is unaware of the state constraint represented by
  Statement~\ref{eqn:axiom-unstab}, it observes the unexpected outcome
  of not having the white cube on the top of the duck after put it
  there, requiring a new plan after step 4. However, the robot learns
  from the unexpected observation and induces the previously unknown
  axiom (from the corresponding decision tree based on the relational
  representation). When asked to achieve the same goal from the same
  initial conditions, the robot computes the following plan:
  \begin{enumerate}
  \item Pick up the duck in step 0;
  \item Put down the duck on the table in step 1;
  \item Pick up the white cube in step 2;
  \item Put down the white cube on the table in step 3;
  \item Pick up the red can in step 4;
  \item Put down the red can on the white cube in step 5.
  \end{enumerate}
  The robot now avoids putting down the white cube (and thus the red
  can) on top of the duck that has irregular surface; executing this
  plan results in the goal being achieved. This example illustrates
  how learning previously unknown state constraints can help agents
  creating better plans. Table~\ref{tab:plans2} shows quantitative
  results, which were obtained by considering multiple different
  scenarios and goals, to further support this conclusion.  }
\end{execexample}

\smallskip
\begin{execexample}\label{exec-example2}[Axiom merging and generalization] \\
  {\rm Consider an agent with the following rule in its knowledge
    base:
  \begin{align}
  \label{eqn:axiom-learnover}
  &\neg occluded(A) \leftarrow \ not~obj\_relation(behind, A, B),
  obj\_relation(above, A, C)
  \end{align}
  This axiom is an over-specification of the axiom encoded by
  Statement~\ref{eqn:axiom-learn}(b); specifically, it contains the
  unnecessary literal $obj\_relation(above, A, C)$. Now, consider the
  situation in which the axiom learning approach has managed to
  extract a correct (ground) version of this axiom encoded by
  Statement~\ref{eqn:axiom-learn}(b). This invokes the axiom merging
  approach described in Section~\ref{sec:arch-dtmerge}, which proceeds
  as follows.
  \begin{itemize}
  \item The robot compares the newly discovered axiom with the
    existing over-specified version (in
    Statement~\ref{eqn:axiom-learnover}) to recognize that they are
    different version of the same axiom.
  \item The two axioms are placed in different ASP programs and tested
    (along with other axioms) in a number of different (artificially
    constructed) scenarios.
  \item The more general (i.e., concise) version of the axiom achieves
    better accuracy in these scenarios and the over-specified version
    is discarded.
  \item Recall that the axiom merging approach attributes an initial
    strength to learned axioms that decreases over time. If this
    strength falls bellow a threshold, the corresponding axiom is
    discarded. This helps retain only the relevant axioms for
    subsequent reasoning.
  \end{itemize}
  This example illustrates the working of
  Algorithm~\ref{alg:learning}, and shows how it leverages the
  existing knowledge to learn and revise axioms.  }
\end{execexample}

%%%%%%%%%%%%%%%%%%%%%%%%%%%%%%%%%%%%%%%%%%%%%%%%%%%%%%%%%%%%%%%%%%%%%%%%%%%%%%%%%%%%
\subsection{Experimental Results}
\label{sec:expres-results}
We next describe and discuss quantitative results corresponding to the
evaluation of the hypotheses described in
Section~\ref{sec:expres-setup}. The first set of experiments was
designed to test the incremental grounding of spatial relations (i.e.,
hypothesis \underline{\textbf{H1}}) as follows, using the setup
procedure from Section~\ref{sec:expres-setup-grounding}, with the
results summarized in Table~\ref{tab:comb}:
\begin{enumerate}
\item Pairs of objects extracted from the training set of the TOSD
  were randomly divided into $10$ subsets.

\item Seven pairs of objects from each subset were used to train the
  MSR-based grounding with human feedback. Each pair represents one of
  the position-based spatial relations under consideration (i.e.,
  \textit{in, left, right, front, behind, above, below}).

\item Seven pairs of objects from each subset labeled with human
  feedback, and $200$ pairs with relations labeled using the QSR-based
  grounding, were used to train the MSR-based grounding.

\item The control node chose either the QSR-based grounding or the
  MSR-based grounding trained using the QSR-based grounding and human
  feedback.
\end{enumerate} 
The three schemes (in Steps 2-4 above) were evaluated on $200$ object
pairs in test scenes of varying complexity. Table~\ref{tab:comb}
indicates that the MSR-based grounding acquired using the QSR-based
grounding makes better use of human feedback than the MSR-based
grounding acquired using just human feedback, which supports
hypothesis \underline{\textbf{H1}}. Note that the same amount of human
feedback is provided with the scheme in Step-2 and the scheme in
Step-3. The difference is that the latter scheme bootstraps off the
generic knowledge encoded in the QSR-based grounding. These results
indicate that performance is improved by using prior knowledge,
experience, and human feedback, and an appropriate representation for
knowledge. Also, the control node-based combination of the two
groundings (scheme in Step-4) provides better accuracy than just using
the QSR-based approach (with accuracy of $70\%$ not shown in
Table~\ref{tab:comb}) or just using the MSR-based approach (scheme in
Step-2). These results also support hypothesis
\underline{\textbf{H1}}.

\begin{table}[tb]
\centering 
%\vspace{-0.5em}
%\hline
\begin{tabular}{|m{1.7cm} | m{2.5cm} | m{3cm} | m{2.5cm}|}\hline
  & \multicolumn{3}{ m{8cm} }{Accuracy of labels over test set of $200$ object pairs} \\ \cline{2-4}
 &&\\[-2ex]
%\hline
Training sets & MSR (feedback) & MSR (QSR + feedback) & Combined model\\[0.5ex]
\hline
&&\\[-2ex]
Sets 1 & 65\% & 77\% & 84\%\\[0.5ex]
%\hline
&&\\[-2ex]
Sets 2 & 82\% & 80\% & 94\%\\[0.5ex]
%\hline
&&\\[-2ex]
Sets 3 & 68\% & 80\% & 85\%\\[0.5ex]
%\hline
&&\\[-2ex]
Sets 4 & 66\% & 83\% & 87\%\\[0.5ex]
%\hline
&&\\[-2ex]
Sets 5 & 65\% & 74\% & 82\%\\[0.5ex]
%\hline
&&\\[-2ex]
Sets 6 & 68\% & 77\% & 86\%\\[0.5ex]
%\hline
&&\\[-2ex]
Sets 7 & 64\% & 87\% & 90\%\\[0.5ex]
%\hline
&&\\[-2ex]
Sets 8 & 64\% & 84\% & 91\%\\[0.5ex]
%\hline
&&\\[-2ex]
Sets 9 & 62\% & 82\% & 87\%\\[0.5ex]
%\hline
&&\\[-2ex]
Sets 10 & 52\% & 72\% & 81\%\\[0.5ex]
\hline
&&\\[-2ex]
Mean & 65\% & 79\% & 87\%\\[0.5ex]
%\hline
&&\\[-2ex]
Std Dev & 7.2\% & 4.6\% & 8.3\% \\ \hline
\end{tabular}
\caption{Comparison of three schemes (1) MSR-based grounding trained with just human feedback; (2) MSR-based grounding trained with 200 pairs labeled by the QSR-based grounding and seven pairs labeled with human feedback; and (3) the use of a control node to choose between the MSR-based grounding trained as in \#2 and QSR-based grounding. The combined model supported by the third scheme provides significantly better performance than the other two schemes.}
%\vspace{-0.75em}
\label{tab:comb}
\end{table} 

\begin{figure}[tb]
 \begin{center}
  \includegraphics[width=0.6\textwidth]{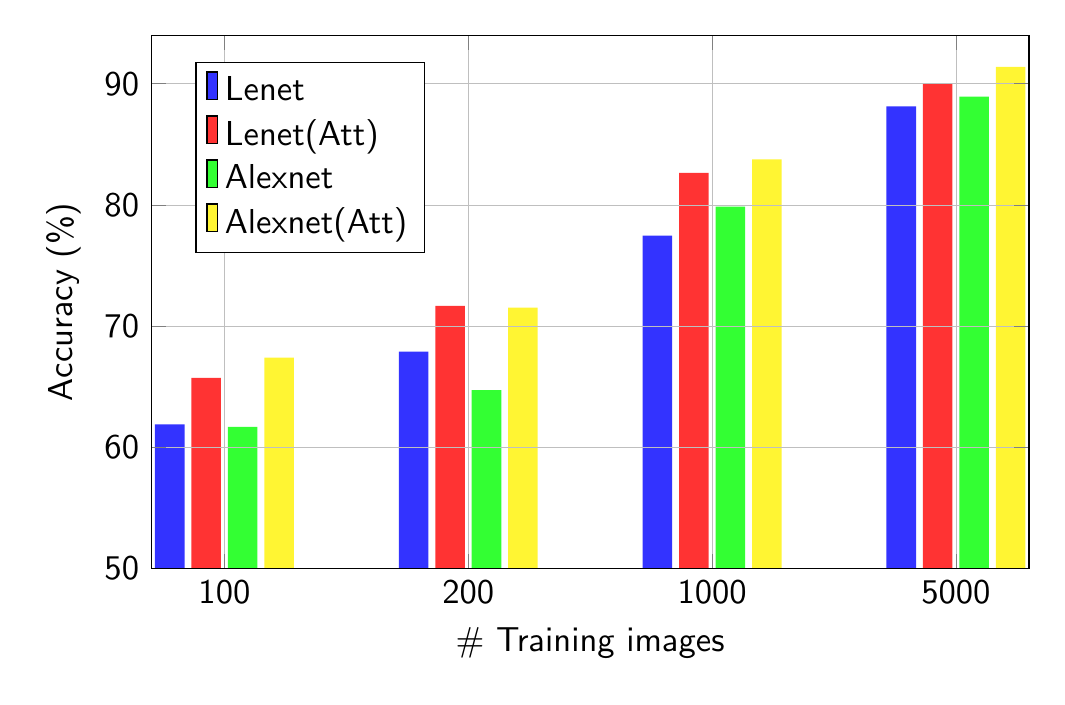}
  \vspace{-2em}
  \caption{Accuracy of Lenet and Alexnet with and without commonsense
    reasoning and attention mechanism. Number of background images
    were $100$. Our architecture improves accuracy in comparison with
    the baselines.}
  \label{fig:CNN_performance}
 \end{center}
 \vspace{-1em}
\end{figure}

The subsequent experiments evaluated the ability of our architecture
to assign occlusion and stability labels to objects and object
structures in images, following the setup described in
Section~\ref{sec:expres-setup-estimate}. Specifically, the second set
of experiments was designed as follows, with results summarized in
Figure~\ref{fig:CNN_performance}:
\begin{enumerate}
\item Training datasets of different sizes ($100$, $200$, $1000$, and
  $5000$ images) were used to train the Lenet and Alexnet networks.
  The remaining images were used to test the learned models. Recall
  that the baseline CNNs do not use the attention mechanism or
  commonsense reasoning; the corresponding results are plotted as
  ``Lenet'' and ``Alexnet'' in Figure~\ref{fig:CNN_performance};
 
\item Another instance of the Lenet and Alexnet networks were trained
  and tested as part of our architecture, i.e., as directed by the
  reasoning module and the attention mechanism module. This training
  and testing considered the same images as in Step-1 but
  automatically identified the relevant ROIs and extracted the
  corresponding data from the second dataset described in
  Section~\ref{sec:expres-setup-estimate}. The corresponding results
  are plotted as ``Lenet(Att)'' and ``Alexnet(Att)'' in
  Figure~\ref{fig:CNN_performance}.
\end{enumerate}

\noindent
Figure~\ref{fig:CNN_performance} indicates that reasoning with
commonsense knowledge and using the attention mechanism to guide deep
learning improves accuracy in comparison with the baselines (based on
deep networks) for the estimation of stability and occlusion. Training
and testing the deep networks with only relevant ROIs of images that
cannot be processed by commonsense reasoning simplifies and directs
the learning process.  This approach helps learn an accurate mapping
between inputs and outputs, resulting in higher accuracy than the
baselines for any given number of training images. The improvement is
more pronounced when the training set is smaller, but there is
improvement at all training dataset sizes considered in our
experiments. These results support hypothesis \underline{\textbf{H2}}.

\begin{figure}[tb]
  \begin{center}
    \begin{subfigure}{0.35\textwidth}
      \includegraphics[width=\textwidth]{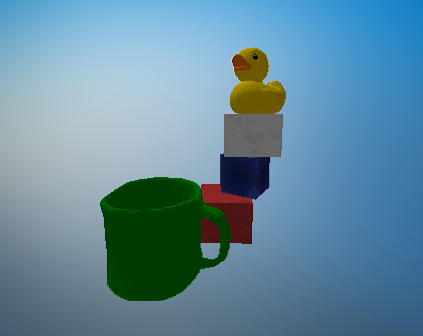}
      \caption{}%Example image of scene.}
      \label{fig:image1}
    \end{subfigure}
    \hspace{0.1in}
    \begin{subfigure}{0.34\textwidth}
      \includegraphics[width=\textwidth]{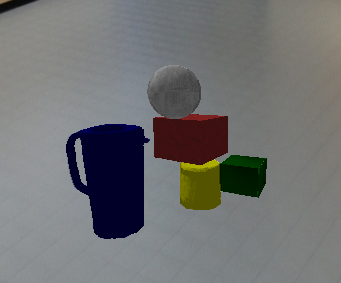}
      \caption{}%Point clouds of scene.}
      \label{fig:image2}
    \end{subfigure}
  \end{center}
  \vspace{-2em}
  \caption{Test images for Lenet and Lenet(Att) architectures: (a)
    both networks detected the occlusion of the red cube by green mug,
    but only the latter correctly estimated the tower's instability;
    and (b) both networks predicted the instability of the tower, but
    only Lenet(Att) detected the obstruction of green cube by yellow
    cylinder.}
  \label{fig:example-image}
  \vspace{-1em}
\end{figure}

Figure~\ref{fig:example-image} shows two specific instances of scenes
in which the inclusion of commonsense reasoning and the attention
mechanism improves performance. In Figure~\ref{fig:image1}, both
\emph{Lenet} and \emph{Lenet(Att)} were able to recognize the
occlusion of the \emph{red cube} caused by the \emph{green mug}, but
only the latter, which uses the attention mechanism and commonsense
reasoning, was able to estimate the instability of the tower. In
Figure~\ref{fig:image2}, both networks correctly predicted the
instability of the tower. However, only \emph{Lenet(Att)} was able to
identify the occlusion of the \emph{green cube} by the \emph{yellow
  can}. The classification errors of baselines architectures are
primarily because a similar example had not been observed during
training---this is a known limitation of deep network architectures.
The attention mechanism eliminates the analysis of unnecessary parts
of images and focuses only on the relevant parts, resulting in a more
targeted network that provides better classification accuracy.  For
these experiments, the CNNs were trained with $1000$ images.
 
\begin{figure}[tb]
  \begin{center}
   \includegraphics[width=0.6\textwidth]{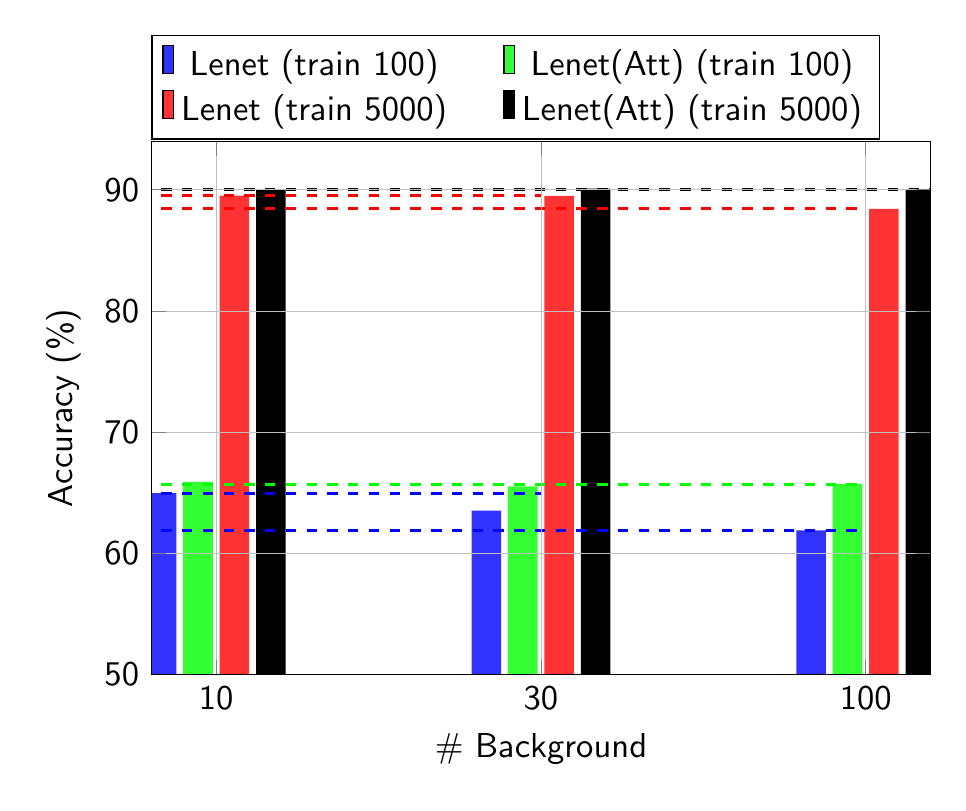}
   \vspace{-2em}
   \caption{Effect of changing the number of backgrounds on the
     accuracy of the Lenet and Lenet(Att) networks for $100$ and
     $5000$ training images. Without the commonsense reasoning and
     attention mechanism, variations in the background influence the
     classification accuracy.}
  \label{fig:bkgrd_effect}
 \end{center}
 \vspace{-1.5em}
\end{figure}

The number of different backgrounds (selected randomly) was fixed at
$100$ for the experimental results in
Figure~\ref{fig:CNN_performance}. The effect of the background on the
observed performance varies depending on the number of training
examples. For instance, we had (on average) one image that used each
background image when the training data had $100$ training samples,
and we had $50$ images per background for the training dataset with
$5000$ training examples. However, in real scenarios, it is unlikely
that we will get a uniform distribution of backgrounds; other factors
such as lighting, viewpoint, and orientation will be different in
different images. To analyze the effect of different backgrounds, we
explored the use of the Lenet architecture with different number of
training examples ($100$ and $5000$) and different number of
backgrounds ($30$, $50$, and $100$). As shown in
Figure~\ref{fig:bkgrd_effect}, varying the background did have an
impact on accuracy, which degraded from $\approx 65\%$ for one
background per $10$ images to $\approx 62\%$ when we had one
background per image (i.e., $100$ backgrounds for $100$ images). The
degradation was smaller, i.e., $\approx 1\%$, for $5000$ training
examples with number of backgrounds varying from $10-100$; however,
for $1000$ backgrounds (one background per five training images) the
accuracy was reduced by $\approx 2\%$.  These results indicate that a
network trained with a larger number of images is less sensitive to
variations in background, and that the inclusion of different
backgrounds has a negative effect on the performance of the baseline
(e.g., Lenet) architecture. On the other hand, with the inclusion of
commonsense reasoning and the attention mechanisms, i.e., with
Lenet(Att), classification accuracy similar over a range of the number
of backgrounds, which indicates that our architecture provides some
robustness to background variations.

\begin{figure}
  \begin{center}
    \includegraphics[width=0.6\textwidth]{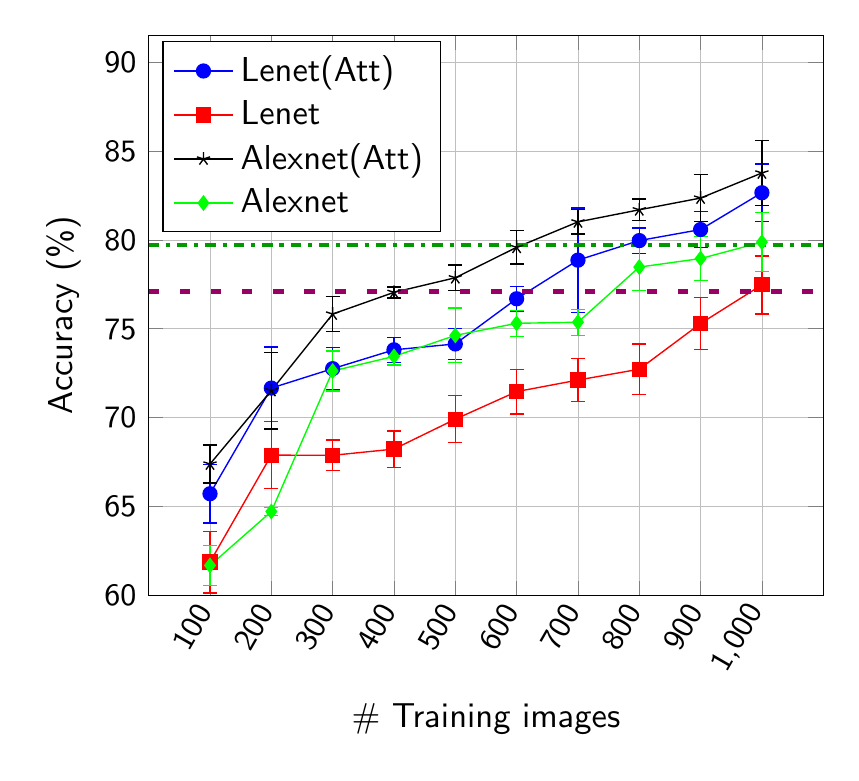}
    \vspace{-2em}
    \caption{Accuracy of Lenet and Alexnet with and without the
      attention mechanism and commonsense reasoning. The number of
      background images was fixed at $100$. Any desired accuracy is
      achieved with a smaller training set when commonsense reasoning
      and the attention mechanism are used.}
    \label{fig:CNN_stddev}
  \end{center}
  \vspace{-1.5em}
\end{figure}

%\noindent
The third set of experiments was designed as follows to evaluate
hypothesis \underline{\textbf{H3}} on the computational effort of
training the deep networks in our architecture, with the corresponding
results summarized in Figure~\ref{fig:CNN_stddev}:
\begin{enumerate}
\item The Lenet and Alexnet networks were trained with training
  datasets containing between $100-1000$ images, in step-sizes of
  $100$. Separate datasets were created for testing. Recall that the
  baseline deep networks do not include commonsense reasoning or the
  attention mechanism;

\item Another instance of the Lenet and Alexnet networks were trained
  and tested as part of our architecture, i.e., as directed by the
  reasoning module and the attention mechanism module. This training
  and testing considered the same images as in Step-1 but
  automatically identified the relevant ROIs and extracted the
  corresponding data from the second dataset described in
  Section~\ref{sec:expres-setup-estimate}. The corresponding results
  are plotted as ``Lenet(Att)'' and ``Alexnet(Att)'' in
  Figure~\ref{fig:CNN_stddev}.
\end{enumerate}

\noindent
Figure~\ref{fig:CNN_stddev} shows that using the attention mechanism
and reasoning with commonsense knowledge helps achieve any desired
level of accuracy with much fewer training examples. The purple dashed
line in Figure~\ref{fig:CNN_stddev} indicates that the baseline Lenet
needs $\approx 1000$ images to reach an accuracy of $77\%$, whereas
our architecture reduces this number to $\approx 600$.  The
experiments indicates a similar difference between the Alexnet and
Alexnet(Att) for $79.5\%$ accuracy---see the dark green dash-dotted
line in Figure~\ref{fig:CNN_stddev}. In other words, the use of
commonsense knowledge helps train deep networks with fewer examples,
reducing both the computational requirements and storage requirements.
These results support hypothesis \underline{\textbf{H3}}.

The fourth set of experiments was designed as follows to evaluate
hypothesis \underline{\textbf{H4}}, i.e., the ability to incrementally
learn previously unknown axioms, with the corresponding results
summarized in Table~\ref{tab:simulated}:
\begin{enumerate}
\item Ten sets of $50$ labeled images were created, as described in
  Section~\ref{sec:expres-setup-estimate};
\item The axiom learning algorithm was trained with each set three
  times, using thresholds of $95\%$ and $70\%$ at the leaf nodes of
  the decision trees. These are the values assigned to threshold
  $th_1$ in the algorithm described in Table~\ref{alg:learning} in
  Section~\ref{sec:arch-dtmerge};
\item The precision and recall were computed for learning previously
  unknown axioms, e.g., Statements~\ref{eqn:axiom-learn}(a),
  ~\ref{eqn:axiom-learn}(b), and~\ref{eqn:axiom-unstab}, but excluding
  defaults and with threshold of $th_1 = 95\%$. The corresponding
  results are summarized as ``unknown (normal)'' in
  Table~\ref{tab:simulated};
\item The precision and recall were computed for learning the unknown
  default statements, e.g., Statement~\ref{eqn:axiom-unstabdef} with
  threshold of $th_1 = 70\%$. The results are summarized as ``unknown
  (default)'' in Table~\ref{tab:simulated}.
\end{enumerate}
In the results summarized in Table~\ref{tab:simulated}, errors were
predominantly variants of the target axioms that were not in the most
generic form, i.e., they had irrelevant literals but were not actually
wrong. The lower precision and recall with defaults is understandable
because it is challenging to distinguish between defaults and their
exceptions. Although we do not describe it here, other studies
indicated that reasoning with commonsense knowledge and decision trees
can also be used to provide relational descriptions as (at least
partial) explanations for the decisions made by the
architecture~\cite{mota:SNCS21}.

\begin{table}[tb]
\normalsize
\centering
\caption{Precision and recall for learning previously unknown domain axioms (normal, default) using decision tree induction. It is possible to learn default constraints, but that also introduces errors.}
\label{tab:simulated}
\begin{tabular}{|c|c|c|}%m{2.5cm}  m{2cm}  m{2cm}}
\hline\noalign{\smallskip}
Axiom type & Precision & Recall\\%[0.5ex]
\noalign{\smallskip}\hline\noalign{\smallskip}
%&\\[-2ex]
Unknown (normal) & 98\% & 100\%\\[1.8ex] \hline
%&\\[-2ex]
Unknown (default) & 78\%  & 62\%\\[0.1ex]
\noalign{\smallskip}\hline
\end{tabular}
\end{table}

\begin{figure}[htb]
  \begin{center}
    \includegraphics[width=0.7\textwidth]{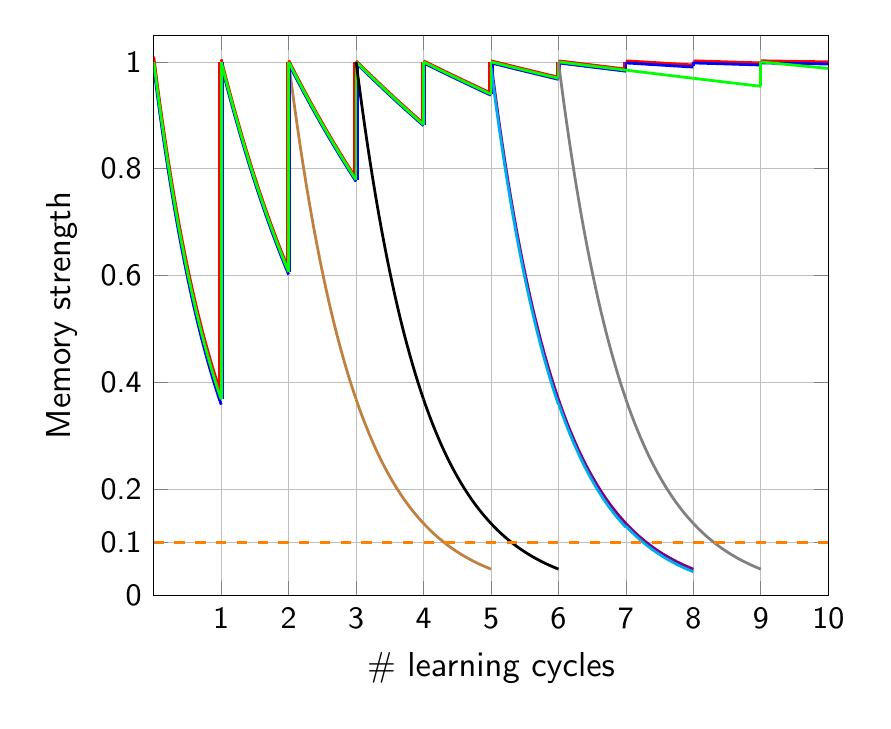}
    \vspace{-2.5em}
    \caption{Evolution of the strengths of learned axioms (excluding
      defaults) over time.}
    \label{fig:normal}
  \end{center}
  \vspace{-1.5em}
\end{figure}

\begin{figure}[tb]
  \begin{center}
    \includegraphics[width=0.7\textwidth]{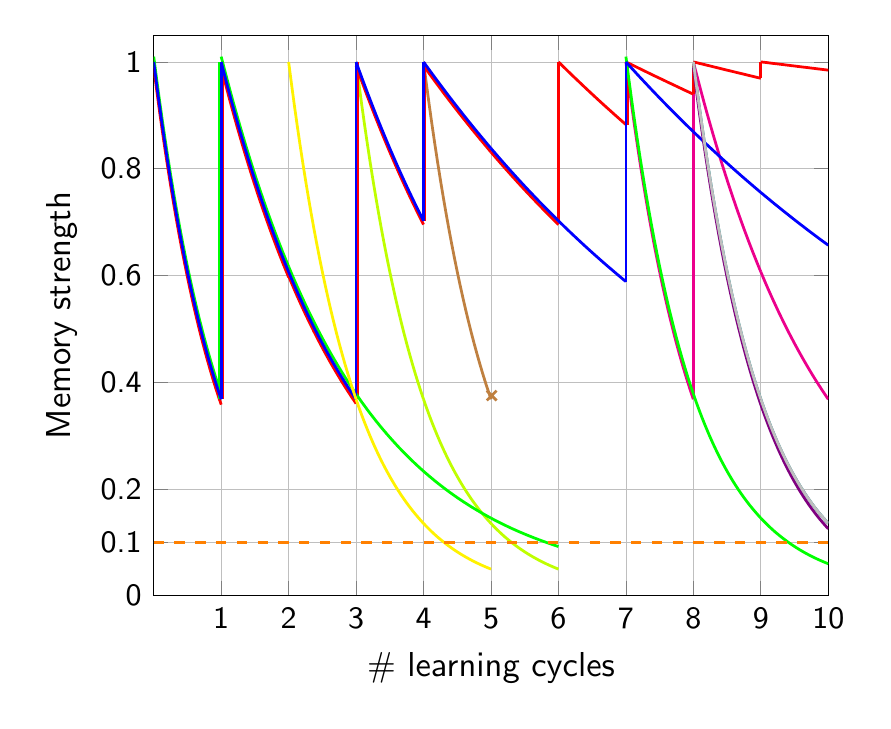}
    \vspace{-2.5em}
    \caption{Evolution of the strengths of learned axioms (including
      defaults) over time.}
    \label{fig:default}
  \end{center}
  \vspace{-1em}
\end{figure}

The fifth set of experiments were designed as follows to evaluate
hypothesis \underline{\textbf{H5}} on revising and merging axioms,
with results summarized in Figures~\ref{fig:normal}
and~\ref{fig:default} for non-default axioms and default axioms
respectively:
\begin{enumerate}
\item Ten sets of $60$ labeled scenes were created, as described in
  Section~\ref{sec:expres-setup-estimate}. Each set was used in one
  run of ensemble learning (see Algorithm~\ref{alg:learning});

\item In each cycle, $50$ images were used for decision tree induction
  and axioms extraction. The other $10$ images supported the choice of
  the best version of similar axioms;

\item The decay factor and axiom strengths were updated, and the
  axioms with strength below $10\%$ were eliminated (orange dashed
  line in the figures below); and
  % The new parameters were applied between two
  % cycles to calculate the new strength; and

\item Steps 2 and 3 were repeated for $10$ learning cycles.
\end{enumerate}

% \begin{table}[htb]
% \normalsize
% \centering
% \caption{Learned axioms (excluding defaults) whose strengths are revised over time, as shown in Figure~\ref{fig:normal}}.
% \begin{tabular}{ l | m{11.5cm}| l}
% %\hline
% Color & Axiom & Discarded?\\
% \hline
% \textcolor{green}{\textbf{------}} & holds(stable(A),I) :- -holds(obj\_relation(above,A,B),I), -has\_surface(A, irregular). & No\\

% \hline
% \textcolor{red}{\textbf{------}}&-holds(occluded(A),I) :- -holds(obj\_relation(behind,A,B),I). & No\\

% \hline
% \textcolor{blue}{\textbf{------}}&-holds(stable(A),I) :- holds(irreg\_below(A),I). & No\\

% \hline
% \textcolor{violet}{\textbf{------}}&-holds(occluded(A),I) :- holds(obj\_relation(above,A,B),I). & Cycle 8\\

% \hline
% \textcolor{brown}{\textbf{------}}&-holds(stable(A),I) :- holds(obj\_relation(above,A,B),I), holds(tower\_height(A,N),I), N$>$4. & Cycle 5\\

% \hline
% \textcolor{cyan}{\textbf{------}}&-holds(stable(A),I) :- holds(small\_base(A),I), holds(tower\_height(A,N),I), N$>$4. & Cycle 8\\

% \hline
% \textbf{------}&holds(stable(A),I) :- -holds(obj\_relation(above,A,B),I), holds(obj\_relation(behind,A,C),I). & Cycle 6\\

% \hline
% \textcolor{gray}{\textbf{------}}&holds(stable(A),I) :- holds(tower
% \_height(A,N),I), N$<=$1. & Cycle 9\\

% \end{tabular}
% \label{tab:normal}

% \end{table}

\begin{table}[htb]
\normalsize
\centering
\caption{Learned axioms (excluding defaults) whose strengths are revised over time, as shown in Figure~\ref{fig:normal}; "obj\_rel" is used as a short form of the relation "obj\_relation", and each fluent $F$ is short form for $holds(F, I)$.}
\begin{tabular}{ | l | m{9.5cm}| l |}
\hline
Color & Axiom & Discarded?\\
\hline
\textcolor{green}{\textbf{------}} & stable(A) :- ~not~ obj\_rel(above,A,B),~ $\lnot$ has\_surface(A, irregular). & No\\

\hline
\textcolor{red}{\textbf{------}}& $\lnot$~occluded(A) :- ~not~ obj\_rel(behind,A,B). & No\\

\hline
\textcolor{blue}{\textbf{------}}&$\lnot$~ stable(A) :- ~irregular\_below(A). & No\\

\hline
\textcolor{violet}{\textbf{------}}&$\lnot$~ occluded(A) :- ~obj\_rel(above,A,B). & Cycle 8\\

\hline
\textcolor{brown}{\textbf{------}}&$\lnot$~ stable(A) :- ~obj\_rel(above,A,B),~ tower\_height(A,N),~ N$>$4. & Cycle 5\\

\hline
\textcolor{cyan}{\textbf{------}}&$\lnot$~ stable(A) :- ~small\_base(A), ~tower\_height(A,N),~ N$>$4. & Cycle 8\\

\hline
\textbf{------}& stable(A) :- ~not~ obj\_rel(above,A,B), ~obj\_rel(behind,A,C). & Cycle 6\\

\hline
\textcolor{gray}{\textbf{------}}& stable(A) :- ~tower\_height(A,N), ~N$<=$1. & Cycle 9\\ \hline
\end{tabular}
\label{tab:normal}
\end{table}

\noindent
Figure \ref{fig:normal} shows how the strength of the eight axioms in
Table~\ref{tab:normal} behaved over $10$ cycles. The top three axioms
are shown in green, red, and blue; they are learned or reinforced in
almost every cycle. Note that the first axiom corresponding to the
green-colored plot is a variant of the axiom in
Statement~\ref{eqn:axiom-learn}(a); it states that any object that is
not above another object and does not have an irregular surface is
stable. Although this axiom was not re-learned in learning cycles $7$
and $8$, it was able to maintain a high value of strength. This is
because it was reinforced in all previous cycles, resulting in a small
decay factor. In contrast, the other five axioms in
Table~\ref{tab:normal} were not learned or used frequently. The
strength of each of these axioms decreased over time and eventually
fell below a threshold, causing the corresponding axiom to be removed.
Figure~\ref{fig:normal} shows that our approach was able to identify
and retain the correct axioms, and Figure~\ref{fig:default} shows
similar results for $10$ learned axioms (that included default
statements) in Table~\ref{tab:default}. Note that among the learned
axioms shown in Table~\ref{tab:default}, those on lines $4$ and $5$
(represented by brown and lime green-colored plots in
Figure~\ref{fig:default}) are different versions of the same axiom.
Specifically, the axiom on line $4$ has an extra (unnecessary) literal
\textit{(holds(obj\_relation(front, A, D), I)}). When the similarity
was identified, the axioms were merged by retaining the more general
version, and the brown-colored plot stops after cycle $5$ in
Figure~\ref{fig:default}. These results support hypothesis
\underline{\textbf{H5}}.

% \begin{table}[tb]
% \normalsize
% \centering
% \caption{Learned axioms (including defaults) whose strengths are revised over time, as shown in Figure~\ref{fig:default}; "obj\_rel" is used as a short form of the relation "obj\_relation".}
% \begin{tabular}{ l | m{10.5cm} | l}

% Color & Axioms & Discarded?\\

% \hline
% \textcolor{red}{\textbf{------}} & -holds(stable(A),I) :- holds(small\_base(A),I). & No\\

% \hline
% \textcolor{blue}{\textbf{------}} & -holds(stable(A),I) :- holds(obj\_relation(above,A,B),I). & No\\

% \hline
% \textcolor{magenta}{\textbf{------}} & -holds(stable(A),I) :- holds(tower\_height(A,N),I), N$>$4. & No\\

% \hline
% \textcolor{brown}{\textbf{------}} & holds(occluded(A),I) :- holds(obj\_relation(behind,A,B),I), -holds(obj\_relation(above,A,C),I), -holds(obj\_relation(front,A,D),I). & Cycle 5\\

% \hline
% \textcolor{lime}{\textbf{------}} & holds(occluded(A),I) :- holds(obj\_relation(behind,A,B),I), -holds(obj\_relation(above,A,C),I). & Cycle 6\\

% \hline
% \textcolor{green}{\textbf{------}} & -holds(occluded(A),I) :- holds(obj\_relation(above,A,B),I). & Cycles 6 and 10\\

% \hline
% \textcolor{teal}{\textbf{------}} & holds(stable(A),I) :- -holds(obj\_relation(above,A,B),I). & No\\

% \hline
% \textcolor{violet}{\textbf{------}} & holds(stable(A),I) :- holds(obj\_relation(behind,A,B),I). & No\\

% \hline
% \textcolor{lightgray}{\textbf{------}} & holds(stable(A),I) :- holds(obj\_relation(front,A,B),I). & No\\

% \hline
% \textcolor{yellow}{\textbf{------}} & holds(stable(A),I) :- -holds(small\_base(A),I). & Cycle 5\\

% \end{tabular}

% \label{tab:default}
% \end{table}

\begin{table}[tb]
\normalsize
\centering
\caption{Learned axioms (including defaults) whose strengths are revised over time, as shown in Figure~\ref{fig:default}; "obj\_rel" is used as a short form of the relation "obj\_relation", and each fluent $F$ is short form for $holds(F, I)$.}
\begin{tabular}{| l | p{11.5cm} | l |}\hline
Color & Axioms & Discarded?\\
\hline
\textcolor{red}{\textbf{------}} & $\lnot$~stable(A) :- ~small\_base(A). & No\\

\hline
\textcolor{blue}{\textbf{------}} & $\lnot$~stable(A) :- ~obj\_rel(above,A,B). & No\\

\hline
\textcolor{magenta}{\textbf{------}} & $\lnot$~stable(A) :- ~tower\_height(A,N),~N$>$4. & No\\

\hline
\textcolor{brown}{\textbf{------}} & occluded(A) :- ~obj\_rel(behind,A,B),~ not~obj\_rel(above,A,C), ~not~obj\_rel(front,A,D). & Cycle 5\\

\hline
\textcolor{lime}{\textbf{------}} & occluded(A) :-~obj\_rel(behind,A,B), ~not~obj\_rel(above, A, C). & Cycle 6\\

\hline
\textcolor{green}{\textbf{------}} & $\lnot$~occluded(A) :- ~obj\_rel(above,A,B). & Cycles 6, 10\\

\hline
\textcolor{teal}{\textbf{------}} & stable(A) :- ~not~obj\_rel(above,A,B). & No\\

\hline
\textcolor{violet}{\textbf{------}} & stable(A) :- ~obj\_rel(behind,A,B). & No\\

\hline
\textcolor{lightgray}{\textbf{------}} & stable(A) :- ~obj\_rel(front,A,B). & No\\

\hline
\textcolor{yellow}{\textbf{------}} & stable(A) :- ~$\lnot$~small\_base(A). & Cycle 5\\ \hline
\end{tabular}
\label{tab:default}
\end{table}

\begin{figure}[tb]
  \begin{center}
    \includegraphics[width=0.5\textwidth]{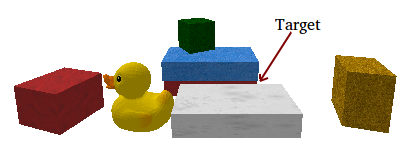}
    \vspace{-1.5em}
    \caption{Illustrative image from the planning experiments with and
      without the learned axioms.}
    \label{fig:image_plan}
  \end{center}
  \vspace{-1.5em}
\end{figure}

\begin{table}[tb]
  \caption{Number of plans and planning time with the learned axioms expressed as a fraction of the values without the learned axioms. Reasoning with the learned axioms improves performance.}
  \label{tab:plans}
  %\vspace{-1em}
  \begin{center}
    \begin{tabular}{|c|c|}
      \hline
      \textbf{Measures} & \textbf{Ratio (with/without)}\\      
      %\cline{2-2}
      %\textbf{Measures} & \textbf{Simulated scenes}\\
      \hline
      Number of plans & 0.33\\
      \hline
%       Optimal & 1 & 1\\
%       \hline
%       Sub-optimal & 0.43 & 0.55\\
%       \hline
%       Incorrect & 0 & 0\\
%       \hline
      Planning time & 0.89\\
      \hline
    \end{tabular}
  \end{center}
  \vspace{-1.5em}
\end{table}

\begin{table}[tb]
  \caption{Number of optimal, sub-optimal, and incorrect plans expressed as a fraction of the total number of plans. Reasoning with the learned axioms improves performance.}
  \label{tab:plans2}
  % \vspace{-1em}
  \begin{center}
    % \scalebox{0.93}{%
    \begin{tabular}{|c|c|c|}
      \hline
     % & \multicolumn{2}{|c|}{\bf Real Scenes} & \multicolumn{2}{|c|}{\bf Simulated Scenes} \\ \cline{2-5}
      \textbf{Plans} & \textbf{Without} & \textbf{With}\\
      \hline
      Optimal &  0.15 & 0.49 \\
      \hline
      Sub-optimal & 0.31 & 0.51 \\
      \hline
      Incorrect & 0.54 & 0 \\
      \hline
    \end{tabular}
    % }
  \end{center}
  \vspace{-1.5em}
\end{table}

\medskip
\noindent
The final set of experiments was designed as follows to evaluate the
second part of hypothesis \textbf{H4}:
\begin{enumerate}
\item Forty initial object configurations were arranged;
  Figure~\ref{fig:image_plan} shows an illustrative example;

\item For each initial state, five goals were randomly chosen and
  encoded in the ASP program. The robot reasoned with the existing
  knowledge to create plans for these $200$ combinations ($40$ initial
  states, five goals);

\item The plans were evaluated in terms of the number of optimal,
  sub-optimal and incorrect plans, and the time taken to compute the
  plan; and

\item Experiments were repeated with and without the learned axioms.
  % with the results for the former expressed as a fraction of those
  % for the latter.
\end{enumerate}
Since the number of plans and planning time vary depending on the
initial conditions and the goal, we conducted paired trials with and
without the learned constraints included in the ASP program used for
reasoning. The initial conditions and goal were identical in each
paired trial, and differed between different paired trials. Then, we
expressed the number of plans and the planning time with the learned
constraints as a fraction of the corresponding values obtained by
reasoning without the learned constraints. The average of these
fractions over all the trials is reported in Table~\ref{tab:plans}. In
addition, we computed the number of optimal, sub-optimal, and
incorrect plans in each trial as a fraction of the total number of
plans in the trial; we did this separately with and without using the
learned axioms for reasoning, and the average over all trials is
summarized in Table~\ref{tab:plans2}. These results indicate that
using the learned axioms for reasoning significantly reduced the
search space, resulting in a much smaller number of plans and a
substantial reduction in the planning time. In addition, when the
robot used the learned axioms for reasoning, it resulted in a much
smaller number of sub-optimal plans and eliminated all incorrect
plans. Also, each such sub-optimal plan was created only when the
corresponding goal could not be achieved without creating an exception
to a default, e.g., stacking an object on a small base. Without the
learned axioms, a larger fraction of the plans are sub-optimal or
incorrect. These results support hypothesis \underline{\textbf{H4}}.

As a specific example of the planning trials, the goal in one trial
was to move the large red box partially hidden behind the white box
and the duck in Figure~\ref{fig:image_plan} such that it is no longer
occluded.  With all the axioms the robot found eight plans (all of
which were correct); however, with some axioms missing, the robot
found as many as $90$ plans, many of which were incorrect. A plan was
considered to be correct if executing it (in simulation) resulted in
the corresponding goal being achieved.

%%%%%%%%%%%%%%%%%%%%%%%%%%%%%%%%%%%%%%%%%%%%%%%%%%%%%%%%%%%%%%%%%%%%%%%%%%%%%%%%%%%%
%%%%%%%%%%%%%%%%%%%%%%%%%%%%%%%%%%%%%%%%%%%%%%%%%%%%%%%%%%%%%%%%%%%%%%%%%%%%%%%%%%%%
\section{Conclusions}
\label{sec:conclusions}
Deep network architectures and algorithms represent the state of the
art for many tasks in robotics and AI.  However, they require large
labeled training datasets and considerable computational resources,
and it is difficult to understand the decisions made by the learned
networks. The architecture described in this paper draws inspiration
from research in cognitive systems to address these limitations.
Instead of focusing of generalizing across domains, our architecture
integrates the complementary principles of deep learning,
non-monotonic logical reasoning with commonsense knowledge, and
decision tree induction of knowledge for reliable and efficient
reasoning and learning for any given domain. The underlying intuition
is that commonsense knowledge is available in almost every application
domain---in fact, some such knowledge is often used implicitly or
explicitly to optimize the parameters of deep networks. Our
architecture seeks to fully exploit this knowledge.  Reasoning with
domain knowledge simplifies learning---the robot only needs to learn
aspects of the domain not already encoded, or encoded incorrectly, by
the existing knowledge. A more accurate mapping is thus learned
between the desired inputs and outputs using a smaller set of labeled
examples and based on fewer attributes. We have experimentally
validated our intuition in the context of estimating the occlusion of
objects and the stability of object structures in simulated and
real-world images. Our architecture improves accuracy, and reduces
storage and computation requirements, especially when large labeled
training datasets are not available.

The architecture described here opens up multiple directions for
future research. First, we will consider more complex scenes with more
objects and clutter, e.g., by including more complex real-world images
from the TOSD dataset similar to the image in Figure~\ref{fig:tosd2},
in our experiments on estimating occlusion and stability. We
hypothesize that our architecture will scale to such scenes because it
reasons with, and learns from, only the relevant information in the
scene. Second, we will further explore the interplay between reasoning
and learning to better understand the operation of deep network
models. We have already shown in other work that the use of relational
logical structures makes it easier to explain the decisions, the
underlying knowledge and beliefs, and the experiences that informed
these beliefs~\cite{mota:SNCS21}. We will now explore the use of
different deep network architectures, and use the corresponding
learned axioms to better understand the behavior of the deep networks.
Note that our architecture's ability to selectively train the deep
networks on a subset of images also simplifies the process of
exploring the behavior of these networks. Third, we will expand the
learning capabilities of our architecture to learn other kinds of
axioms. Recall that we have only discussed the learning of state
constraints in this paper, although the underlying representation
supports other kinds of axioms as well. Other work in our group has
demonstrated the use of relational reinforcement learning, inductive
learning, and limited human feedback for learning different types of
axioms~\cite{mota:aaaisymp20,mohan:ACS18}. Incorporating the ability
to learn different axioms in our architecture will significantly
enhance our ability to reason accurately and better explore the
behavior of the deep networks. Furthermore, we will explore the use of
this architecture on robots assisting humans in complex domains,
reasoning with relevant information at different
resolutions~\cite{mohan:JAIR19}, and providing relational descriptions
as explanations for their decisions and beliefs during reasoning and
learning.

%%%%%%%%%%%%%%%%%%%%%%%%%%%%%%%%%%%%%%%%%%%%%%%%%%%%%%%%%%%%%%%%%%%%%%%%%%%%%%%%%%%%
%%%%%%%%%%%%%%%%%%%%%%%%%%%%%%%%%%%%%%%%%%%%%%%%%%%%%%%%%%%%%%%%%%%%%%%%%%%%%%%%%%%%
%\section*{Acknowledgements}

%%%%%%%%%%%%%%%%%%%%%%%%%%%%%%%%%%%%%%%%%%%%%%%%%%%%%%%%%%%%%%%%%%%%%%%%%%%%%%%%%%%%
%%%%%%%%%%%%%%%%%%%%%%%%%%%%%%%%%%%%%%%%%%%%%%%%%%%%%%%%%%%%%%%%%%%%%%%%%%%%%%%%%%%%
\section*{Declarations}
%Please consider the following declarations.
%\medskip
\noindent
\textit{Funding:} This work was supported in part by the Asian Office
of Aerospace Research and Development award FA2386-16-1-4071. All
opinions and conclusions described in this paper are those of the
authors.

\medskip
\noindent
\emph{Conflict of interest/Competing interests:} Not
applicable.

\medskip
\noindent
\emph{Availability of data and material:} The entire dataset of images
used in our experiments is $\ge 100MB$ and point cloud data is $\ge
3GB$ in size. A link to most of the data is provided with our code
repository (see below); we will make all the data available online
after review.

\medskip
\noindent
\emph{Code availability:} Please
see~\cite{code-results}.

\medskip
\noindent
\emph{Author contributions:} TM and MS contributed to the design of
the architecture and algorithms; TM implemented the algorithms; TM and
MS designed the experiments; TM conducted the experiments and gathered
experimental results; TM and MS analyzed the results; MS and TM wrote
the paper.

%%%%%%%%%%%%%%%%%%%%%%%%%%%%%%%%%%%%%%%%%%%%%%%%%%%%%%%%%%%%%%%%%%%%%%%%%%%%%%%%%%%%
%%%%%%%%%%%%%%%%%%%%%%%%%%%%%%%%%%%%%%%%%%%%%%%%%%%%%%%%%%%%%%%%%%%%%%%%%%%%%%%%%%%%

% BibTeX users please use one of
%\bibliographystyle{spbasic}      % basic style, author-year citations
%\bibliographystyle{spmpsci}      % mathematics and physical sciences
%\bibliographystyle{spphys}       % APS-like style for physics

\bibliographystyle{plain}      % basic style, author-year citations
\bibliography{references}   % name your BibTeX data base

% Non-BibTeX users please use
%\begin{thebibliography}{}
%
% and use \bibitem to create references. Consult the Instructions
% for authors for reference list style.
%
%\bibitem{RefJ}
% Format for Journal Reference
%Author, Article title, Journal, Volume, page numbers (year)
% Format for books
%\bibitem{RefB}
%Author, Book title, page numbers. Publisher, place (year)
% etc
%\end{thebibliography}

\end{document}